\documentclass[5p]{elsarticle}

\usepackage{lineno}
\usepackage{hyperref}
\usepackage{amsmath}
\usepackage{xcolor}
\usepackage{layouts}
\usepackage{graphicx}
\usepackage{adjustbox}
\usepackage{booktabs}
\usepackage{subfig}

\usepackage[normalem]{ulem}
\usepackage{color}

\newcommand{\rem}[1]{}

\biboptions{authoryear}

\makeatletter
\def\ps@pprintTitle{%
 \let\@oddhead\@empty
 \let\@evenhead\@empty
 \def\@oddfoot{\textit{\hfill\today}}%
 \let\@evenfoot\@oddfoot}
\makeatother

\begin{document}
\setlength{\emergencystretch}{3em}
\begin{frontmatter}

\title{A high-resolution canopy height model of the Earth}

\author[1]{Nico Lang\corref{cor1}}
\author[2]{Walter Jetz}
\author[1]{Konrad Schindler}
\author[1,3]{Jan Dirk Wegner}

\address[1]{EcoVision Lab, Photogrammetry and Remote Sensing, ETH Z\"urich, Switzerland}
\address[2]{Department of Ecology and Evolutionary Biology, Yale
University, USA}
\address[3]{Institute for Computational Science, University of Zurich, Switzerland}
\cortext[cor1]{Corresponding author: nico.lang@geod.baug.ethz.ch}

\begin{abstract}
The worldwide variation in vegetation height is fundamental to the global carbon cycle \citep{dubayah2020global,essd-2021-386} and central to the functioning of ecosystems and their biodiversity \citep{migliavacca2021three, skidmore2021priority}. Geospatially explicit and, ideally, highly resolved information is required to manage terrestrial ecosystems \citep{felipe2018multiple}, mitigate climate change, and prevent biodiversity loss \citep{skidmore2021priority}. 
Here, we present the first global, wall-to-wall canopy height map at 10$\,$m ground sampling distance for the year 2020.
No single data source meets these requirements: dedicated space missions like GEDI \citep{dubayah2020global} deliver sparse height data, with unprecedented coverage, whereas optical satellite images like Sentinel-2 offer dense observations globally, but cannot directly measure vertical structures.
By fusing GEDI with Sentinel-2, we have developed a probabilistic deep learning model to retrieve canopy height from Sentinel-2 images anywhere on Earth, and to quantify the uncertainty in these estimates.
The presented approach reduces the saturation effect commonly encountered when estimating canopy height from satellite images \citep{potapov2021mapping, healey2020highly}, allowing to resolve tall canopies with likely high carbon stocks \citep{jucker2017allometric,DUNCANSON2022112845}. According to our map, only 5\% of the global landmass is covered by trees taller than 30$\,$m. 
Such data play an important role for conservation, e.g., we find that only 34\% of these tall canopies are located within protected areas \citep{protectedplanet2021}. Our model enables consistent, uncertainty-informed worldwide mapping and supports an ongoing monitoring to detect change and inform decision making. The approach can serve ongoing efforts in forest conservation \citep{UN2017forest}, and has the potential to foster advances in climate \citep{jucker2018canopy,de2019global}, carbon \citep{DUNCANSON2022112845,dubayah2020global}, and biodiversity modelling \citep{tuanmu2015global,skidmore2021priority,tuia2022perspectives}.
\end{abstract}


\end{frontmatter}

\section{Introduction}\label{intro}
As our society depends on a multitude of terrestrial ecosystem services \citep{manning2018redefining}, the conservation of the Earth's forests has become a priority on the global political agenda \citep{UN2017forest}.
To ensure sustainable development through biodiversity conservation and climate change mitigation, the United Nations have formulated global forest goals that include maintaining and enhancing global carbon stocks and increasing forest cover by 3\% between 2017 and 2030 \citep{UN2017forest}.
Yet global demand for commodities is driving deforestation, impeding progress towards these ambitious goals \citep{hoang2021mapping}. 
Earth observation and satellite remote sensing play a key role in this context, as they provide the data to monitor the quality of forested area at global scale \citep{hansen2013high}. 
However, to measure progress in terms of carbon and biodiversity conservation, novel approaches are needed that go beyond detecting forest cover and can provide consistent information about morphological traits predictive of carbon stock and biodiversity \citep{skidmore2021priority}, at global scale. One key vegetation characteristic is canopy height \citep{skidmore2021priority, jucker2017allometric}.

In this work, we describe a deep learning framework to map canopy top height globally with high resolution, using publicly available optical satellite images as input.
We deploy that model to compute the first global canopy top height product with 10$\,$m ground sampling distance, based on Sentinel-2 optical images for the year 2020. That global map and underlying source code and models, are made publicly available to support conservation efforts as well as science in disciplines such as climate, carbon, and biodiversity modelling.\footnote{The map can be explored interactively in this \href{https://nlang.users.earthengine.app/view/global-canopy-height-2020}{browser application}.}

Mapping canopy height in a consistent fashion at global scale is key to understand \textit{terrestrial ecosystem functions}, which are dominated by vegetation height and vegetation structure \citep{migliavacca2021three}.
Canopy top height is an important indicator of biomass and the associated, global \textit{aboveground carbon stock} \citep{DUNCANSON2022112845}. At high spatial resolution, canopy height models (CHM) directly characterize \textit{habitat heterogeneity} \citep{tuanmu2015global}, which is why canopy height has been ranked as a high-priority \textit{biodiversity variable} to be observed from space \citep{skidmore2021priority}.
Furthermore, forests buffer microclimate temperatures under the canopy \citep{de2019global}. While it has been shown that in the tropics higher canopies provide a stronger dampening effect on microclimate extremes \citep{jucker2018canopy}, targeted studies are needed to see if such relationships also hold true at global scale \citep{de2019global}. Thus, a homogeneous high-resolution CHM has the potential to advance the modelling of \textit{climate impact} on terrestrial ecosystems, and may assist forest management to bolster microclimate buffering, as a mitigation service to protect biodiversity under a warmer climate \citep{de2019global}.

Given forests' central relevance to life on our planet, several new space missions have been developed to measure vegetation structure and biomass. A key mission is NASA's GEDI campaign, which has been collecting full-waveform LIDAR data explicitly for the purpose of measuring vertical forest structure globally, between 51.6$^\circ$ north and south \citep{dubayah2020global}. 
GEDI has unique potential to advance our understanding of the global carbon stock, but its geographical range, and also its spatial and temporal resolutions, are limited. The mission length, initially set to two years, does not allow for continuous forest monitoring into the future. Moreover, GEDI is a sampling mission expected to cover at most 4\% of the land surface. By design, the collected samples sparsely cover the surface of the Earth, which restricts the resolution of wall-to-wall mission products to a 1$\,$km grid \citep{dubayah2020global}. 
In contrast, satellite missions such as Sentinel-2 or Landsat, which have been designed for a broader range of Earth observation needs, deliver freely accessible archives of optical images that are not as tailored to vegetation structure, but offer longer-term global coverage at high spatial and temporal resolution. Sensor fusion between GEDI and multi-spectral optical imagery has the potential to overcome the limitations of each individual data source \citep{valbuena2020standardizing}.

However, estimating forest characteristics like canopy height or biomass from optical images is a challenging task \citep{gibbs2007monitoring}, as the physical relationships between spectral signatures and vertical forest structure are complex and not well understood \citep{rodriguez2017quantifying}.
Given the vast amount of data collected by the GEDI mission, we circumvent this lack of mechanistic understanding by harnessing supervised machine learning, in particular end-to-end deep learning. From millions of data examples, our model learns to extract patterns and features from raw satellite images which are predictive of high-resolution vegetation structure.
By fusing GEDI observations with Sentinel-2 images, our approach enhances the spatial and temporal resolution of the CHM, and extends its geographic range to the sub-arctic and arctic regions outside of GEDI's coverage.
While retrieval of vegetation parameters with deep learning has been demonstrated regionally, and up to country scale \citep{lang2019country, becker2021country, lang2021high, rodriguez2021mapping}, we scale it up and process the entire global landmass. This step presents a technical challenge but is crucial to enable operational deployment and ensure consistent, globally homogeneous data.

\section{Deep learning approach}

\begin{figure*}[th!]
    \centering
    \includegraphics[width=1.0\textwidth]{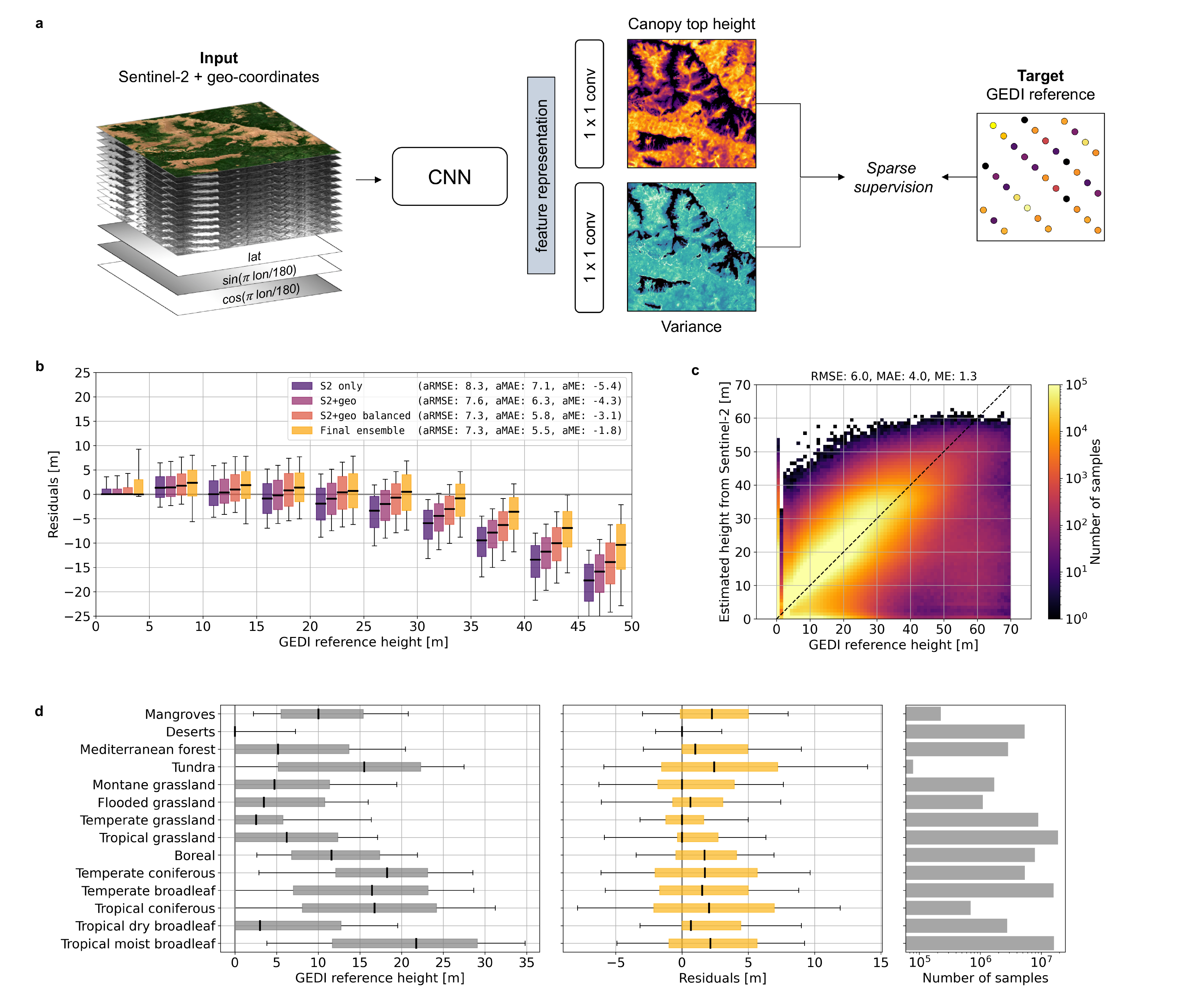}
    \caption{Model overview and global model evaluation on held-out GEDI reference data.
    a) Illustration of the model training process with sparse supervision from GEDI LIDAR. The CNN takes the Sentinel-2 image and encoded geographical coordinates as an input to estimate dense canopy top height as well as its predictive uncertainty (variance).
    b) Residual analysis w.r.t.\ canopy height intervals, and ablation study of model components. Negative residuals indicate that estimates are lower than reference values.
    c) Confusion plot for the final model ensemble, showing good agreement between predictions from Sentinel-2 and GEDI reference.
    d) Biome-level analysis of final ensemble estimates: GEDI reference height, residuals, and number of samples per biome.
    }
    \label{fig:modelling}
\end{figure*}

Deep learning is revolutionizing fields ranging from medicine \citep{jumper2021highly} to weather forecasting \citep{ravuri2021skilful} and has great potential to advance environmental monitoring \citep{reichstein2019deep,tuia2022perspectives}, but its application to global remote sensing is technically challenging due to the large data volume \citep{reichstein2019deep,kattenborn2021review}. 
Cloud platforms like Google Earth Engine \citep{gorelick2017google} simplify the analysis of satellite data but provide a limited set of traditional machine learning tools that depend on manual feature design. To use them, one must sacrifice some flexibility in terms of methods, in return for easy access to large data archives and compute power.
In particular, canopy height estimation with existing standard tools tends to struggle with the underestimation of tall canopies, as the height estimates saturate around 25 to 30$\,$m \citep{hansen2016mapping, potapov2021mapping, healey2020highly}. This is a fairly severe limitation in regions dominated by tall canopies, like tropical forests, and deteriorates downstream carbon stock estimation, since tall trees have especially high biomass \citep{jucker2017allometric}.
A further restriction of prior large-scale CHM projects is that they rely on local calibration, which hampers their use in locations without nearby reference data \citep{potapov2021mapping, healey2020highly}. Technically, existing mapping schemes aggregate reflectance data over time, but perform pure pixel-to-pixel mapping without regard to local context and image texture.

Here, we extend previous regional deep learning methods \citep{lang2019country, becker2021country, lang2021high} to a global scale. These methods have been shown to mitigate the saturation of tall canopies by exploiting texture, while not depending on temporal features \citep{lang2019country}.
In more detail, our approach employs an ensemble \citep{lakshminarayanan2017simple} of deep, (fully) convolutional neural networks (CNN), each of which takes as input a Sentinel-2 optical image and transforms it into a dense canopy height map with the same ground sampling distance (GSD) of 10$\,$m \citep{lang2019country}. See Fig.~\ref{fig:modelling}a.
Our unified, global model is trained with sparse supervision, using reference heights at globally distributed GEDI footprints, derived from the raw waveforms \citep{lang2022global}. 
A dataset of 600 million samples is constructed by extracting Sentinel-2 image patches of 15$\times$15 pixels around every GEDI footprint acquired between April and August in 2019 or 2020. The sparse GEDI data is rastered to the Sentinel-2 10-meter grid by setting the pixel corresponding to the center of each GEDI footprint to the associated footprint height.
In this way, during model training one can optimize the loss function w.r.t.\ the model parameters only at valid reference pixels (Fig.~\ref{fig:modelling}a); while during map generation the CNN model will nevertheless output a height prediction for every input pixel.
To evaluate the model globally, we split the collected dataset at the level of Sentinel-2 tiles. I.e., of the 100~km$\times$100~km regions defined by the Sentinel-2 tiling 20\% are held out for validation, the remaining 80\% are used to train the model (the validation regions and the associated estimation errors are shown in Extended Data Fig.~\ref{fig:map_errors}).

An important goal of our work are low estimation errors for tall vegetation. To that end we extend the CNN model in three ways (Fig.~\ref{fig:modelling}b).
First, we equip the model with the ability to learn geographical priors, by feeding it geographical coordinates (in a suitable cyclic encoding) as additional input channels (Fig.~\ref{fig:modelling}a). Second, we employ a fine-tuning strategy where the sample loss is re-weighted inversely proportional to the sample frequency per 1-meter height interval, so as to counter the bias in the reference data towards low canopies (which reflects the long-tailed world-wide height distribution, where low vegetation dominates, and high values are comparatively rare). Finally, we train an ensemble of CNNs and aggregate predictions from repeated observations of the same location, which reduces the underestimation of tall canopies even further. The combination of all three measures yields the best performance. The average root mean square error (aRMSE) of the height estimates, balanced across all 5-meter height intervals, is 7.3$\,$m, and the average mean error (aME, i.e. bias) is -1.8$\,$m (Fig.~\ref{fig:modelling}b).
The overall RMSE over all validation samples (without height balancing) is 6.0$\,$m, with a bias of 1.3$\,$m (Fig.~\ref{fig:modelling}c). The latter is due to a slight overestimation of low canopy heights, and is the price we pay for improving the performance on rare tall canopies (Fig.~\ref{fig:modelling}b).

A biome-level analysis based on the 14 biomes defined by The Nature Conservancy\footnote{\href{https://www.nature.org/}{www.nature.org} (2022-04-02)} shows how the bias varies across biomes (Fig.~\ref{fig:modelling}d, Extended Data Fig.~\ref{fig:biome_confusion}). The model is able to correctly identify bare soil in deserts with zero height, with marginal error and no bias. The bias is also low in montane, temperate, and tropical grasslands, as well as in Mediterranean and tropical dry broadleaf forest, but higher in flooded grasslands. The most severe overestimation, on average $\approx$2.5$\,$m is observed for mangroves, tundra and tropical coniferous forests.
The highest spread of residuals is observed in tropical and temperate coniferous forests as well as in the tundra, where we note that the latter is significantly underrepresented in the dataset, as GEDI's range does not extend beyond 51.6$^\circ$ north. Furthermore, the GEDI reference data in these tundra regions (southern part of Kamchatka Mountain and Forest Tundra \& Trans-Baikal Bald Mountain Tundra) appears rather noisy and contaminated with a significant number of outliers (Extended Data Fig.~\ref{fig:biome_confusion}). 

We additionally report an evaluation of our final model against independent reference data from NASA's LVIS airborne LIDAR campaigns \citep{blair2004processing} (Extended Data Fig.~\ref{fig:lvis_eval}), which were designed to deliver canopy top heights comparable to GEDI \citep{dubayah2020global}. 
Across all 12 available LVIS areas, our model yields an RMSE of 7.8$\,$m and a ME of 0.6$\,$m, i.e., the overall RMSE is slightly higher than the one for held-out GEDI samples, whereas the bias is lower.
When evaluating only on the areas outside (north of) the GEDI range, we obtain 4.7$\,$m RMSE and 1.6$\,$m ME, whereas only the remaining areas within the GEDI range yield 8.8$\,$m RMSE and 0.2$\,$m ME (recall that the magnitude of the error correlates with the height).
In other words, our estimates agree well with LVIS even outside of the geographical range of the training data. In the appendix \ref{sec:comparison_umd_eth}, we additionally compare against another global canopy height product based on Landsat image composites \citep{potapov2021mapping}.

\section{Modelling predictive uncertainty}

While deep learning models often exhibit high predictive skill and produce estimates with low overall error, the uncertainty of those estimates is typically not known, or unrealistically low (i.e., the model is over-confident) \citep{guo2017calibration}.
But reliable uncertainty quantification is crucial to inform downstream investigations and decisions based on the map content \citep{DUNCANSON2022112845}, e.g., it can indicate which estimates are too uncertain and should be disregarded \citep{lang2022global}.
To afford users of our CHM a trustworthy, spatially explicit estimate of the map's uncertainty, we integrate probabilistic deep learning techniques. These methods are specifically designed to quantify also the predictive uncertainty, taking into account on the one hand the inevitable noise in the input data; and on the other hand the uncertainty of the model parameters, resulting from the lack of reference data for certain conditions \citep{kendall2017uncertainties}.
In particular, we independently train five deep CNNs that have identical structure, but are initialized with different random 
weights. The spread of the predictions made by such a model ensemble \citep{lakshminarayanan2017simple} for the same pixel is an effective way to estimate model uncertainty (a.k.a.\ epistemic uncertainty), even with small ensemble size \citep{ovadia2019can}. 
Each individual CNN is trained by maximizing the Gaussian likelihood, rather than minimizing the more widely used squared error. Consequently, each CNN outputs not only a point-estimate per pixel, but also an associated variance that quantifies the uncertainty of that estimate (a.k.a.\ its aleatoric uncertainty) \citep{kendall2017uncertainties}.

During inference, we process images from ten different dates (satellite overpasses) within a year at every location to obtain full coverage, and exploit redundancy for pixels with multiple cloud-free observations. Each image is processed with a randomly selected CNN within the ensemble, which reduces computational overhead and can be interpreted as natural test-time augmentation, known to improve the calibration of uncertainty estimates with deep ensembles \citep{ashukha2020pitfalls}.

Finally, we use the estimated aleatoric uncertainties to merge redundant predictions from different imaging dates, by weighted averaging proportional to the inverse variance.
While inverse variance weighting is known to yield the lowest expected error \citep{Strutz16}, we observe a deterioration of the uncertainty calibration for low values (\textless4~m standard deviation). 
We also note that uncertainty calibration varies per biome (Extended Data Fig.~\ref{fig:biome_calibration}), so it may be advisable to re-calibrate in post-processing depending on the intended application and region of interest.
\begin{figure*}[tb]
    \centering
    \includegraphics[width=1.0\textwidth]{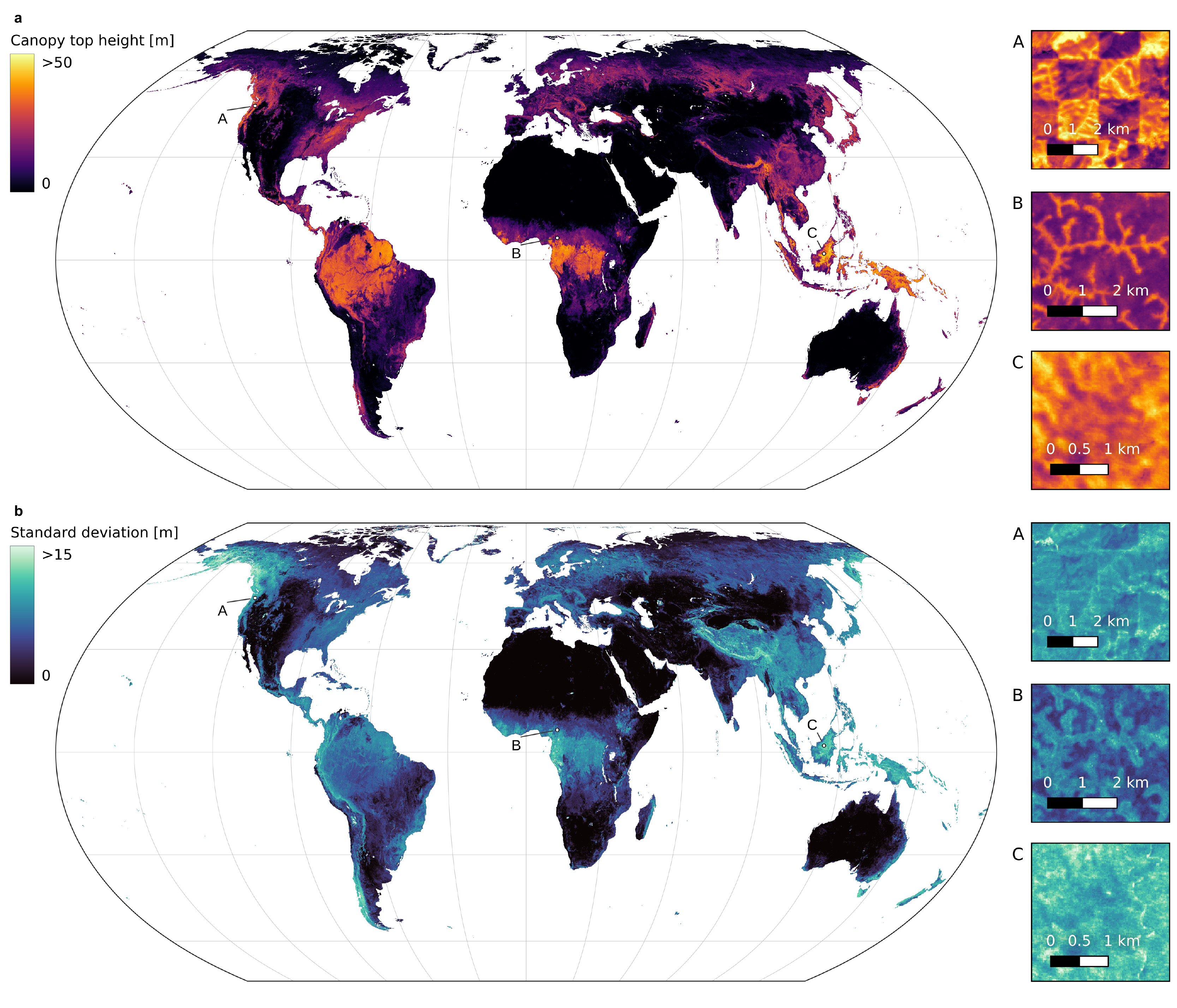}
    \caption{Global canopy height map for the year 2020, visualized in Equal-Earth projection. The underlying data product, estimated from Sentinel-2 imagery, has 10$\,$m ground sampling distance. a) Canopy top height. b) Predictive standard deviation of canopy top height estimates.}
    \label{fig:global_maps}
\end{figure*}
Despite these observations, the estimated predictive uncertainty correlates well with the empirical estimation error and can therefore be used to filter out inaccurate predictions, thus lowering the overall error at the cost of reduced completeness (Extended Data Fig.~\ref{fig:filtering}). E.g., by filtering out the 20\% most uncertain canopy height estimates, overall RMSE is reduced by 13\% (from 6.0$\,$m to 5.2$\,$m) and the bias is reduced by 23\% (from 1.3$\,$m to 1.0$\,$m).

\section{Global canopy height mapping}

\begin{figure*}[tb]
    \centering
    \includegraphics[width=1.0\textwidth]{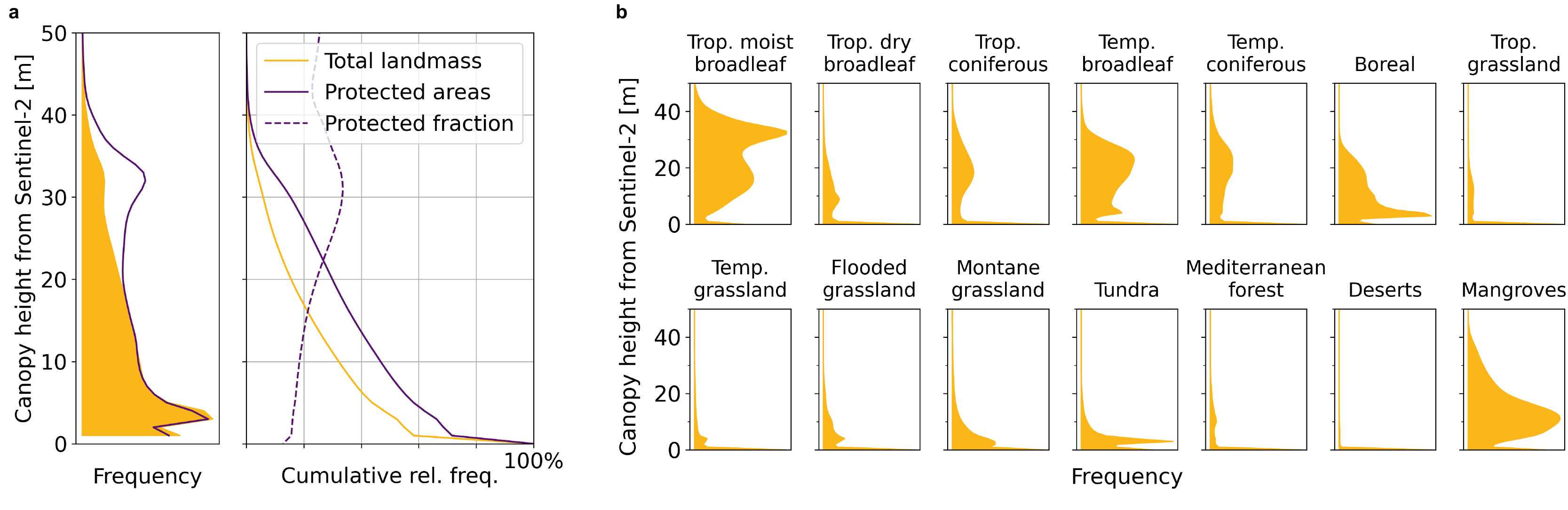}
    \caption{Global canopy height distributions of the entire landmass, protected areas, and biomes. 
    a) Frequency distribution and cumulative distribution for the entire global landmass and within protected areas (according to WDPA \citep{protectedplanet2021}), and fraction of vegetation above a certain height that is protected. 
    b) Biome-level frequency distribution of canopy heights according to 14 terrestrial ecosystems defined by The Nature Conservancy. Urban areas and croplands (based on ESA World Cover \citep{zanaga_daniele_2021_5571936}) have been excluded.}
    \label{fig:global_statistics}
\end{figure*}

The model has been deployed globally on the Sentinel-2 image archive for the year 2020 to produce a global map of canopy top height.
To cover the global landmass ($\approx$1.3$\times10^{12}$ pixels at the GSD of Sentinel-2), a total of $\approx$160 terabyte of Sentinel-2 image data are selected for processing. This required $\approx$27,000 GPU-hours ($\approx$3 GPU-years) of computation time, parallelized on a high-performance cluster to obtain the global map in 10 days real time.

The new wall-to-wall canopy height product at 10$\,$m GSD makes it possible to gain insights into the fine-grained distribution of canopy heights anywhere on Earth, as well as the associated uncertainty (Fig.~\ref{fig:global_maps}). 
Three example locations (A-C in Fig.~\ref{fig:global_maps}) demonstrate the level of canopy detail that the map reveals, ranging from harvesting patterns from forestry in Washington State, US (A), through gallery forests along permanent rivers and ground water in the forest-savanna of Northern Cameroon (B), to dense tropical broadleaf forest in, Borneo, Malaysia (C).

Also at large scale, the predictive uncertainty is positively correlated with the estimated canopy height  (Fig.~\ref{fig:global_maps}b). 
Still, some regions with comparatively low canopy height, like Alaska, northwestern Canada, and Tibet exhibit high predictive uncertainty. The two former lie outside of the GEDI coverage, so the higher uncertainty is likely due to local characteristics that the model has not encountered during training. The latter indicates that, also within the GEDI range, there are environments that are more challenging for the model, e.g., due to globally rare ecosystem characteristics not encountered elsewhere, or frequent cloud cover. 

Our new dataset enables a full, worldwide assessment of coverage of the global landmass with vegetation.
Doing this for a range of thresholds recovers an estimate of the global canopy height distribution (for the year 2020, Fig.~\ref{fig:global_statistics}a and Extended Data Fig.~\ref{fig:protected_areas}a).
We find that an area of 53.6$\times10^6$~km$^2$ (41\% of the global landmass) is covered by vegetation with \textgreater5$\,$m canopy height, 39.6$\times10^6$~km$^2$ (30\%) by vegetation \textgreater10$\,$m, and 6.7$\times10^6$~km$^2$ (5\%) by vegetation \textgreater30$\,$m. See Fig.~\ref{fig:global_statistics}b.

We see that protected areas (according to the World Database on Protected Areas, WDPA \citep{protectedplanet2021}) tend to contain higher vegetation compared to the global average (Fig.~\ref{fig:global_statistics}a). Furthermore, 34\% of all canopies \textgreater30$\,$m fall into protected areas (Fig.~\ref{fig:global_statistics}a). 
See Extended Data Fig.~\ref{fig:protected_areas}b for examples of protected areas that show good agreement with mapped canopy height patterns.
This analysis highlights the relevance of the new dataset for ecological and conservation purposes. For instance, canopy height and its spatial homogeneity can serve as an ecological indicator to identify forest areas with high integrity and conservation value. That task requires both dense area coverage at reasonable resolution, and a high saturation level to locate very tall vegetation.

Finally, our new map makes it possible to analyze the exhaustive distributions of canopy heights at the biome level, revealing characteristic frequency distributions and trends within different types of terrestrial ecosystems (Fig.~\ref{fig:global_statistics}b). While, for instance, the canopy heights of tropical moist broadleaf forests follow a bimodal distribution with a mode \textgreater30$\,$m, mangroves have a unimodal distribution with a large spread and heights ranging up to \textgreater40$\,$m. Notably, our model has learned to predict reasonable canopy heights in tundra regions, despite scarce and noisy reference data for that ecosystem.

\section{Discussion}\label{sec:discussion}

\begin{figure*}[tb]
    \centering
    \includegraphics[width=\textwidth]{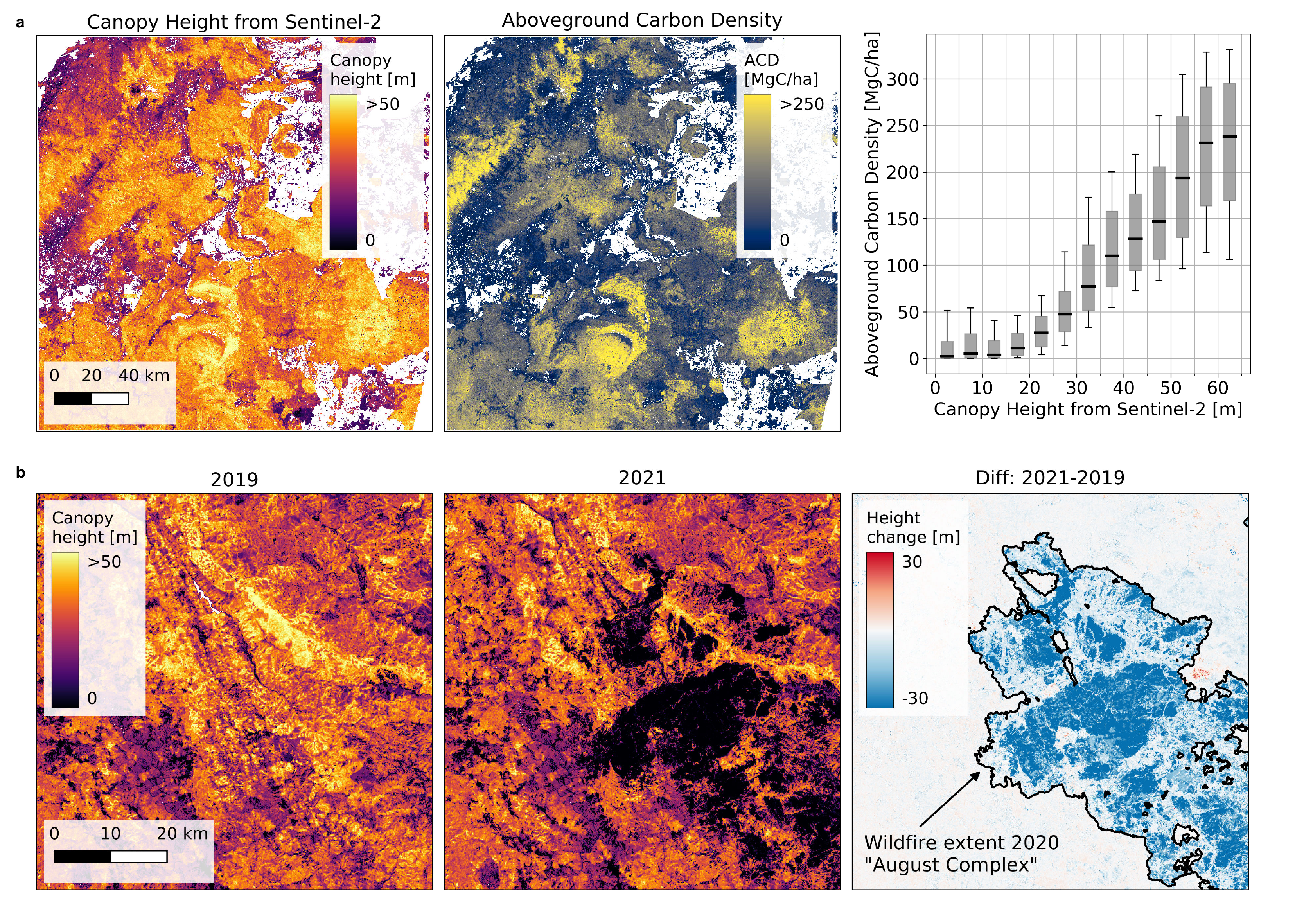}
    \caption{Examples for potential applications. 
    a) Biomass and carbon stock mapping. In Sabah, northern Borneo, canopy height estimated from Sentinel-2 optical images correlates strongly with aboveground carbon density measured by a targeted airborne LIDAR campaign \citep{asner2018mapped}
    (Lat: 5.3812, Lon: 117.0748).
    b) Monitoring environmental damages. In 2020, wildfires destroyed large areas of forests in northern California. The difference between annual canopy height maps is in good agreement with the wildfire extent mapped by the California Department of Forestry and Fire Protection (Lat: 40.1342, Lon: -123.5201).}
    \label{fig:applications}
\end{figure*}

There are at least two major downstream applications that the new high-resolution canopy height dataset can help to advance at global scale, namely biomass and biodiversity modelling. Furthermore, our model can support monitoring of forest disturbances.
Canopy top height is a key indicator to study the global aboveground carbon stock stored in form of biomass \citep{DUNCANSON2022112845}. 
On a local scale, we compare our canopy height map with dense aboveground carbon density data \citep{asner_gregory_p_2021_4549461} that was produced by a targeted airborne LIDAR campaign in Sabah, northern Borneo \citep{asner2018mapped} (Fig.~\ref{fig:applications}a). We observe that for natural tropical forests, the spatial patterns agree well and that our canopy height estimates from Sentinel-2 are predictive of carbon density even in tropical regions, with canopy heights up to 65$\,$m. Notably, the relationship between carbon and canopy height is sensitive up to $\approx$60$\,$m.

We further demonstrate that our model can be deployed annually to map canopy height change over time, e.g., to derive changes in carbon stock and estimate carbon emissions caused by global land-use changes, at present mainly deforestation \citep{essd-2021-386}.
Annual canopy height maps are computed for a region in northern California where wildfires have destroyed large areas in 2020 (Fig.~\ref{fig:applications}b). Our automated change map corresponds well with the mapped fire extent from the California Department of Forestry and Fire Protection\footnote{\href{https://www.frap.fire.ca.gov}{www.frap.fire.ca.gov} (2022-04-02)}, while at the same time the annual maps are consistent in areas not affected by the fires, where no changes are expected.
Such high-resolution change data may potentially help to reduce the high uncertainty of emission estimates that are reported in the annual Global Carbon Budget \citep{essd-2021-386}.
It is worth mentioning that the presented approach yields reliable estimates based on single cloud-free Sentinel-2 images. Thus, its potential for monitoring changes in canopy height is not limited to annual maps, but to the availability of cloud-free images, that are taken at least every 5 days globally. This high update frequency makes it relevant for e.g. real-time deforestation alert systems, even in regions with frequent cloud cover.

A second line of potential applications includes biodiversity modelling, as the high spatial resolution of the canopy height model brings about the possibility to characterize habitat structure and vegetation heterogeneity, known to promote a number of ecosystem services \citep{felipe2018multiple} and to be predictive of biodiversity \citep{macarthur1961bird,tews2004animal}.
The relationship between heterogeneity and species diversity is founded in niche theory \citep{macarthur1961bird,tews2004animal}, which suggests that heterogeneous areas provide more ecological niches for different species to co-exist. Our dense map makes it possible to study second-order homogeneity \citep{tuanmu2015global} (which is not easily possible with sparse data like GEDI), and down to a length scale of 10 to 20$\,$m. 

Technically, our wall-to-wall map makes it a lot easier for scientists to intersect sparse sample data, e.g., field plots, with canopy height.
To make full use of scarce field data in biomass or biodiversity research, dense complementary maps are a lot more useful: when pairing sparse field samples with other sparsely sampled data, the chances of finding enough overlap are exceedingly low; whereas pairing them with low-resolution maps risks biases due to the large scale difference and associated spatial averaging.  

We note that, despite the nominal 10$\,$m ground sampling distance of our global map, the effective spatial resolution at which individual vegetation features can be identified is lower. As a consequence of the GEDI reference data used to train the model, each map pixel effectively indicates the largest canopy top height within a GEDI footprint ($\approx$25$\,$m diameter) centred at the pixel.
Two subtle reasons further impact the effective resolution, compared to a map learned from dense, local reference data (e.g., airborne LIDAR \citep{lang2019country}): the sparse supervision means that the model never explicitly sees high-frequency variations of the canopy height between adjacent pixels; and small misalignments caused by the geolocation uncertainty of the GEDI footprints \citep{dubayah2020global,roy2021impact} introduce further noise.

While at present we do not see this as severely limiting the utility of the map, in the future one could consider extending the method with techniques for guided super-resolution \citep{lutio2019guided}, in order to better preserve small features visible in the raw Sentinel-2 images, like canopy gaps.

Regarding map quality, besides minor artifacts in regions with persistent cloud cover, we observe tiling artifacts at high latitudes in the northern hemisphere.
The systematic inconsistencies at tile borders point at degradation of the absolute height estimates, possibly caused by a lack of training data for particular, locally unique vegetation structures.
Interestingly, it appears that a significant part of these errors are constant offsets between the tiles. Further investigations are needed to ascertain how strongly differential quantities like homogeneity are affected by those artifacts.

\section{Conclusion}\label{sec:conclusion}
We have developed a deep learning method for canopy height retrieval from Sentinel-2 optical images. That method has made it possible to produce the first globally consistent, wall-to-wall canopy top height map of the entire Earth, with 10$\,$m ground sampling distance.
Besides Sentinel-2, the GEDI LIDAR mission also plays a key role, as the source of sparse, but extremely well-distributed reference data at global scale. Compared to previous work that maps canopy height at global scales \citep{potapov2021mapping}, our model substantially reduces the overall underestimation bias for tall canopies. 
Our model does not require local calibration, and can therefore be deployed anywhere on Earth, including regions outside the GEDI coverage.
Moreover, it also delivers globally homogeneous estimates for the predictive uncertainties of the retrieved canopy heights.
As our method, once trained, only needs image data, maps can be updated annually, opening up the possibility to track the progress of commitments, made under the United Nation's global forest goals, to enhance carbon stock and forest cover by 2030 \citep{UN2017forest}.
At the same time, the longer the GEDI mission will collect on-orbit data, the denser the reference data for our approach will become, which can be expected to diminish the predictive uncertainty of its predictions.

As a possible future extension, our model could be extended to map other vegetation characteristics \citep{becker2021country} at global scale. In particular, it appears feasible to densely map biomass, by retraining with GEDI L4A biomass data \citep{DUNCANSON2022112845}, or by adding additional data from planned future space missions \citep{le2018biomass}.

While deep learning technology for remote sensing is continuously being refined by focusing on improved performance at regional scale, its operational utility has been limited by the fact that it often could not be scaled up to global coverage.
Our work demonstrates that, if one has a way of collecting globally well-distributed reference data, modern deep learning can be scaled up and employed for global vegetation analysis from satellite images.
We hope that our example may serve as a foundation on which new, scalable approaches can be built that harness the potential of deep learning for global environmental monitoring; and that it inspires machine learning researchers to contribute to environmental and conservation challenges.

\clearpage
\section{Methods}\label{sec11}

\subsection{Data}
This work builds on data from two ongoing space missions: the Copernicus Sentinel-2 mission operated by the European Space Agency (ESA) and NASA' Global Ecosystem Dynamics Investigation (GEDI). 
The Sentinel-2 multispectral sensor delivers optical images covering the global landmass with a revisit time of at most 5 days on the equator. We use the atmospherically corrected L2A product, consisting of 12 bands ranging from the Visible (RGB) and Near Infra-Red (NIR) to the Short Wave Infra-Red (SWIR). While the RGB and NIR bands have 10$\,$m GSD, the other bands have a 20$\,$m or 60$\,$m GSD. For our purposes all bands are upsampled with cubic interpolation to obtain a data-cube with 10$\,$m ground sampling distance.  
The GEDI mission is a space-based full-waveform LIDAR mounted on the International Space Station, and measures vertical forest structure at sample locations with 25$\,$m footprint, distributed between 51.6$^\circ$ north and south. We use footprint-level canopy top height data derived from these waveforms \citep{lang2022global,lang_nico_2021_5704852} as sparse reference data. The canopy top height is defined as RH98, the relative height at which 98\% of the energy has been returned, and was derived from GEDI L1B waveforms collected between April and August in the years 2019 and 2020.

To train the deep learning model, a global training dataset has been constructed within the GEDI range, by combining the GEDI data and the Sentinel-2 imagery. For every Sentinel-2 tile, we select the image with the least cloud coverage between May and September 2020. Image patches of 15$\times$15 pixels (i.e., 150$\,$m $\times$150$\,$m on the ground) are extracted from these images at every GEDI footprint location. The GEDI data is rastered to the Sentinel-2 pixel grid by setting the canopy height reference value of the pixel that corresponds to the center of the GEDI footprint. Locations for which the image patch is cloudy or snow-covered are filtered out from the dataset. To correct noise injected by the geolocation uncertainty of the GEDI footprints, we use the Sentinel-2 L2A scene classification and assign 0$\,$m canopy height to footprints located in the categories "not vegetated" or "water". This procedure also addresses the slight positive bias due to slope in the GEDI reference data \citep{lang2022global}. Overall, the resulting dataset contains 600$\times10^6$ samples globally distributed within the GEDI range. All samples located within 20\% of the Sentinel-2 tiles in that range (each 100$\,\times\,$100$\,$km) are set aside as validation data (See Extended Data Fig.~\ref{fig:map_errors}). 

A second evaluation is carried out w.r.t.\ canopy top heights independently derived from airborne LIDAR data acquired by NASA's Land, Vegetation, and Ice Sensor (LVIS). LVIS is a large-footprint full-waveform LIDAR, from which the LVIS L2 height metric RH98 is rastered to the Sentinel-2 10$\,$m grid. Locations where LVIS data is available are visualized in Extended Data Fig.~\ref{fig:lvis_eval}.

\subsection{Deep fully convolutional neural network}
Our model is based on the fully convolutional neural network architecture proposed in \citep{lang2019country}. The architecture employs a series of residual blocks with separable convolutions \citep{chollet2017xception}, without any downsampling within the network. The sequence of learnable 3$\times$3 convolutional filters is able to extract not only spectral, but also textural features. To speed-\ up the model for deployment at global scale we reduce its size, setting the number of blocks to 8 and the number of filters per block to 256. 
This speeds up the forward pass by a factor of $\approx$17 compared to the original, larger model. In our tests, the smaller version did not cause higher errors in an early phase of training. When trained long enough, a larger model with higher capacity may be able to reach lower prediction errors, but the higher computational cost of inference would limit its utility for repeated, operational use.
The model takes the 12 bands from Sentinel-2 L2A product and the cyclic encoded geographical coordinates per pixel as input, for a total of 15 input channels. Its output are two channels with the same spatial dimension as the input, one for the mean height and one for its variance. See Fig.~\ref{fig:modelling}. Since the architecture is fully convolutional, it can process arbitrarily sized input image patches, which is useful when deploying it at large scale. 

\subsection{Model training with sparse supervision}
Formally, canopy height retrieval is a pixel-wise regression task. We train the regression model end-to-end in supervised fashion, which means that the model learns to transform raw image data into  spectral and textural features predictive of canopy height, and there is no need to manually design feature extractors. We train the convolutional neural network with sparse supervision, i.e., by selectively minimizing the loss (Eq.~\ref{eq:gaussian_NLL}) only at pixel locations for which there is a GEDI reference value. Before feeding the 15-channel data cube to the CNN, each channel is normalized to be standard normal, using the channel statistics from the training set. The reference canopy heights are normalized in the same way, a common preprocessing step to improve the numerical stability of the training. Each neural network is trained for 5,000,000 iterations with a batch size of 64, using the ADAM optimizer \citep{KingmaB14}. The base learning rate is initially set to 0.0001 and then reduced by factor 0.1 after 2'000'000 iterations and again after 3,500,000 iterations. This schedule was found to stabilizes the uncertainty estimation.

\subsection{Modelling the predictive uncertainty}
Modelling uncertainty in deep neural networks is challenging due to their strong non-linearity, but crucial to build trustworthy models.
The approach followed in this work accounts for two sources of uncertainty, namely the data (aleatoric) and the model (epistemic) uncertainty \citep{kendall2017uncertainties}. 
The uncertainty in the data, resulting from noise in the input and reference data, is modeled by minimizing the Gaussian negative log likelihood (Eq.~\ref{eq:gaussian_NLL}) as a loss function \citep{kendall2017uncertainties}. This corresponds to independently representing the model output at every pixel $i$ as a conditional Gaussian probability distribution over possible canopy heights, given the input data, and estimating the mean $\hat{\mu}$ and variance $\hat{\sigma}^2$ of that distribution.
\begin{equation}
    \centering
    \mathcal{L}_{NLL} = \frac{1}{N} \sum_{i=1}^{N} \frac{ \left ( \hat{\mu}(x_i) -y_i   \right )^2}{2 \hat{\sigma}^2(x_i) } + \frac{1}{2}\log \hat{\sigma}^2(x_i).
    \label{eq:gaussian_NLL}
\end{equation}

To account for the model uncertainty, which in high-capacity neural network models can be interpreted as the model's lack of knowledge about patterns not adequately represented in the training data, we train an ensemble \citep{lakshminarayanan2017simple} of 5 CNNs from scratch, i.e., each time starting the training from a different randomly initialized set of model weights (learnable parameters). 
At inference time we process images from $T$ different acquisition dates (here $T=10$) for every location, to obtain full coverage and to exploit redundancy in the case of repeated cloud-free observations of a pixel. Each image is processed with one CNN picked randomly from the ensemble. This procedure incurs no additional computational cost compared to processing all images with the same CNN. It can be interpreted as a natural variant of test-time augmentation, which has been demonstrated to improve the calibration of uncertainty estimates from deep ensembles in the domain of computer vision \citep{ashukha2020pitfalls}.
Finally, the per-image estimates are merged into a final map by averaging with inverse-variance weighting (Eq.~\ref{eq:mean_prediction}). If the variance estimates of all ensemble members are well calibrated, this results in the lowest expected error \citep{Strutz16}. Thus, the variance of the final per-pixel estimate is computed with the weighted version of the law of total variance (Eq.~\ref{eq:predictive_var}) \citep{kendall2017uncertainties}. For readability we omit the pixel index $i$.

\begin{equation}
     \hat{p}_t =  \frac{1 / \hat{\sigma}_{t}^2}{\sum_{j=1}^{T} 1 / \hat{\sigma}_{j}^2},
    \label{eq:inv_var_weight}
\end{equation}
\begin{equation}
     \hat{y} = \sum_{t=1}^{T} \hat{p}_t \hat{\mu}_{t},
    \label{eq:mean_prediction}
\end{equation}
\begin{equation}
     \mathrm{Var}(\hat{y}) = \sum_{t=1}^{T} \hat{p}_t \hat{\mu}_{t}^2 - \left( \sum_{t=1}^{T} \hat{p}_t \hat{\mu}_{t}  \right)^2 + \sum_{t=1}^{T} \hat{p}_t \hat{\sigma}_{t}^2,
    \label{eq:predictive_var}
\end{equation}

\subsection{Correction for imbalanced height distribution}

We find that the underestimation bias on tall canopies is partially due to the imbalanced distribution of reference labels (and canopy heights overall), where large height values occur rarely.
To mitigate it, we fine-tune the converged model with a cost-sensitive version of the loss function. 
A softened version of inverse sample-frequency weighting is used to re-weight the influence of individual samples on the loss (Eq.~\ref{eq:sqrt_freq}).
To establish the frequency distribution of the continuous canopy height values in the training, we bin them into 1$\,$m height intervals, and in each of the resulting $K$ bins count the number of samples $N_k$. 
Empirically, for our task the moderated reweighting with the square root of the inverse frequency works better (leaving all other hyper-parameters unchanged).
Moreover, we do not fine-tune all model parameters, but only the final regression layer that computes mean height (see Fig.~\ref{fig:modelling}a). 
We observe that the uncertainty calibration is preserved when fine-tuning only the regression weights for the mean ("S2+geo balanced: mean" in Extended Data Fig.~\ref{fig:calibration}), whereas fine-tuning also the regression of the variance decalibrates the uncertainty estimation ("S2+geo balanced: mean\&var"). 
The fine-tuning is run for 750,000 iterations per network.

\begin{equation}
    \centering
    q_i = \frac{\sqrt{1/N_{k, i\in k}}}{\sum_{j=1}^{K} \sqrt{1/N_j}}
    \label{eq:sqrt_freq}
\end{equation}

\subsection{Evaluation metrics}

Several metrics are employed to measure prediction performance: the root mean square error (RMSE, Eq.~\ref{eq:RMSE}) of the predicted heights, their mean absolute error (MAE, Eq.~\ref{eq:MAE}), and their mean error (ME, Eq.~\ref{eq:ME}). The latter quantifies systematic height bias, where a negative ME indicates underestimation, i.e., predictions that are systematically lower than the reference values.
\begin{equation}
    \centering
    \mathrm{RMSE} = \sqrt{ \frac{1}{N} \sum_{i=1}^{N} \left ( \hat{y}_i - y_i   \right )^2 }
    \label{eq:RMSE}
\end{equation}
\begin{equation}
    \centering
    \mathrm{ME} = \frac{1}{N} \sum_{i=1}^{N} \left ( \hat{y}_i - y_i \right )
    \label{eq:ME}
\end{equation}
\begin{equation}
    \centering
    \mathrm{MAE} = \frac{1}{N} \sum_{i=1}^{N} \mid \hat{y}_i - y_i \mid
    \label{eq:MAE}
\end{equation}

We also report balanced versions of these metrics, where the respective error is computed separately in each 5$\,$m height interval and then averaged across all intervals. They are abbreviated as aRMSE, aMAE, and aME (Fig.~\ref{fig:modelling}b).

For the estimated predictive uncertainties, there are by definition no reference values. A common scheme to evaluate their calibration  is to produce calibration plots \citep{guo2017calibration, laves2020well} that show how well the uncertainties correlate with the empirical error. As this correlation holds only in expectation, both the uncertainties and the empirical errors at the test samples must be binned into $K$ equally sized intervals. In each bin $B_k$, the average of the predicted uncertainties is then compared against the actual average deviation between the predicted height and the reference data.
Based on the calibration plots, it is further possible to derive a scalar error metric for the uncertainty calibration, the uncertainty calibration error (UCE) (Eq.~\ref{eq:UCE}) \citep{laves2020well}. Again, we additionally report a balanced version, the average uncertainty calibration error (AUCE) (Eq.~\ref{eq:AUCE}), where each bin has the same weight independent of the number $N_k$ of samples in it.

\begin{equation}
    \centering
    \mathrm{UCE} = \sum_{k=1}^{K} \frac{N_k}{N}  \mid err(B_k) - uncert(B_k) \mid
    \label{eq:UCE}
\end{equation}

\begin{equation}
    \centering
    \mathrm{AUCE} = \frac{1}{K} \sum_{k=1}^{K}  \mid err(B_k) - uncert(B_k) \mid
    \label{eq:AUCE}
\end{equation}
In our case $err(B_k)$ represents the RMSE of the samples falling into bin $B_k$, and the bin uncertainty $uncert(B_k)$ is defined as the root mean variance (RMV):
\begin{equation}
    \centering
    \mathrm{RMV} = \sqrt{\frac{1}{N_k} \sum_{i \in B_k} \hat{u}_i},
    \label{eq:RMV}
\end{equation}
with $\hat{u}_i = \hat{\sigma}_i^2$ when evaluating the calibration of a single CNN, and $\hat{u}_i = \mathrm{Var}(\hat{y}_i)$ when evaluating the calibration of the ensemble. We refer to the RMV as the predictive standard deviation in our calibration plots (Extended Data Fig.~\ref{fig:calibration}, Extended Data Fig.~\ref{fig:biome_calibration}).

\subsection{Global map computation}

Sentinel-2 imagery is organized in 100$\,$km$\times$100$\,$km tiles, a total of 18,011 tiles cover the entire landmass of the Earth, excluding Antarctica. However, depending on the ground tracks of the satellites, some tiles are covered by multiple orbits, whereas in general no more than 2 orbits are need to get full coverage. To optimize computational overhead, we select the relevant orbits per tile by using those with the smallest number of empty pixels, according to the metadata. For every tile and relevant orbit the 10 images with least cloud cover between May and September 2020 are selected for processing.

While it only takes $\approx$2 minutes to process a single image tile with the CNN on a GeForce RTX 2080 Ti GPU, downloading the image from the Amazon Web Service S3 bucket takes about 1 minute, and loading the data into memory takes about 4 minutes. 
To process a full tile with all 10 images per orbit takes between 1 and 2.5 hours, depending on the number of relevant orbits (1 or 2).

We only apply minimal post-processing and mask out built-up areas, snow, ice and permanent water bodies according to the ESA World Cover classification \citep{zanaga_daniele_2021_5571936}, setting their canopy height to "no data" (value: 255).
The canopy height product is released in the form of 3$^\circ\times\,$3$^\circ$ tiles in geographic longitude/latitude, following the format of the recent ESA World Cover product. This choice shall simplify the integration of our map into existing workflows, as well as the intersection of the two products.
Note that the statistics in the present paper were not computed from those tiles, but in Gall-Peters orthographic equal-area projection with 10$\,$m GSD, for exact correspondence between pixel counts and surface areas.

\subsection{Energy and carbon emission footprint}
The presented map has been computed on a GPU cluster located in Switzerland. Carbon accounting for electricity is a complex endeavour, due to differences how electricity is produced and distributed. To put the power consumption needed to produce global maps with our method into context, we estimate carbon emissions for two scenarios, where the computation is run on AWS in two different locations, EU (Stockholm) and US East (Ohio).
With $\approx$250$\,$W to power one of our GPUs, we get a total energy consumption of 250$\,$W$\times$27000$\,$h = 6750$\,$kWh for the global map. 
The conversion to emissions highly depends on the carbon efficiency of the local power grid. For EU (Stockholm) we obtain an estimated 338$\,$kg CO$_2$-equivalent, whereas for US East (Ohio) we obtain 3848$\,$kg CO$_2$-equivalent, a difference by a factor \textgreater10.
While the former is comparable to driving an average car from Los Angeles to San Francisco and back (1,360$\,$km), the latter corresponds to a round trip from Seattle (US) to San José, Costa Rica (15,500$\,$km).
These estimates were conducted using the \href{https://mlco2.github.io/impact#compute}{Machine Learning Impact calculator} \citep{lacoste2019quantifying}.
For the carbon footprint of the current map (not produced with AWS), we estimate $\approx$729$\,$kgCO$_2$eq, using an average of 108$\,$gCO$_2$eq/kWh for Switzerland, as reported for the year 2017 \citep{rudisuli2022decarbonization}.

\section*{Data and code availability}
The global canopy height map for 2020 will be released for download and will be available in the Google Earth Engine.
All links, source code, and the trained models used to generate the map will be released upon publication via the project page: \href{https://langnico.github.io/globalcanopyheight/}{langnico.github.io/globalcanopyheight/}.
The global map can be explored interactively in this browser application: \href{https://nlang.users.earthengine.app/view/global-canopy-height-2020}{nlang.users.earthengine.app/view/global-canopy-height-2020}.

\section*{Acknowledgments}
The project received funding from Barry Callebaut Sourcing AG, as part of a Research Project Agreement.
The Sentinel-2 data access for the computation of the global map was funded by AWS Cloud credits for research, courtesy of Peter Gehler.
We thank Yanina Sica for sharing a rastered version of the WDPA data. 
We thank Noel Gorelick and Simon Ilyushchenko for providing the resources to make our global map available on the Google Earth Engine.
We greatly appreciate the open data policies of the ESA Copernicus program, the NASA GEDI mission, and the LVIS project.

\newpage
\bibliography{sn-bibliography}

\clearpage
\onecolumn
\appendix

\setcounter{section}{0}
\setcounter{figure}{0}
\setcounter{table}{0}

\section{Extended data figures}

\begin{figure*}
    \centering
    \subfloat[]{{\includegraphics[width=0.9\textwidth]{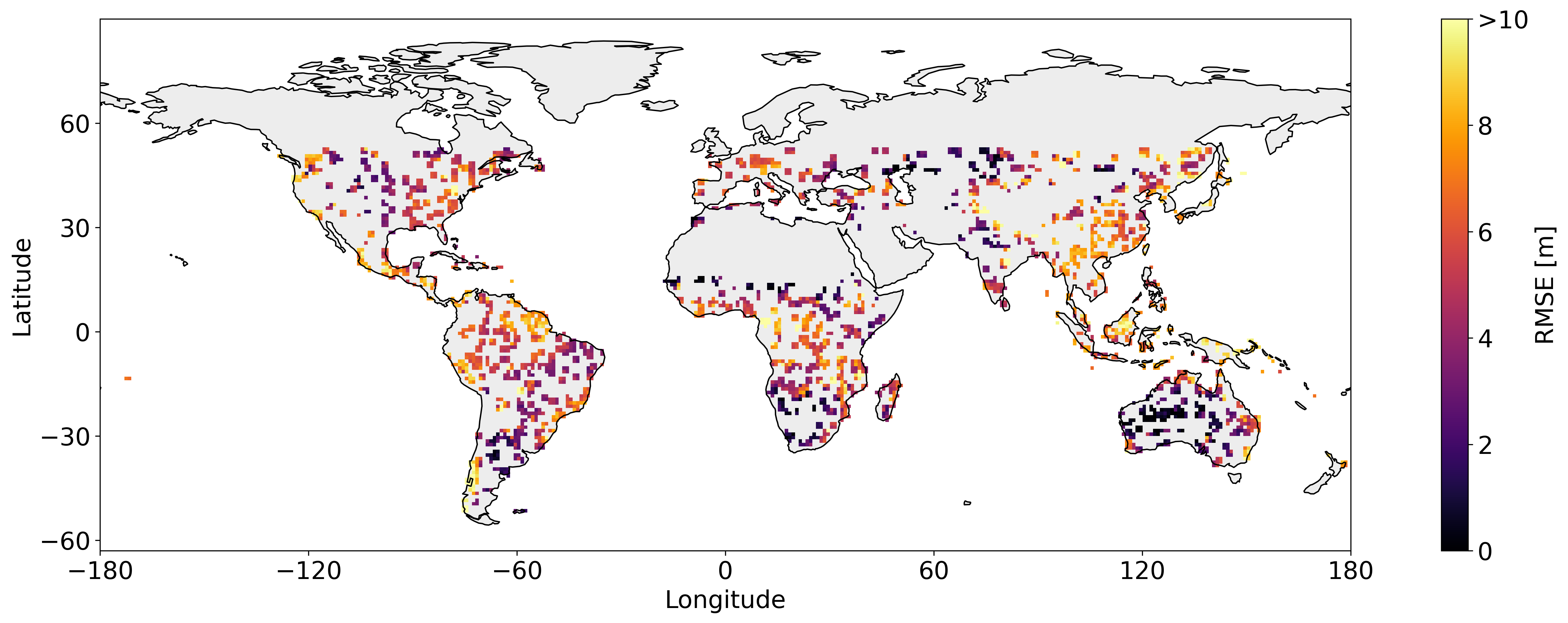} }}%
    
    \subfloat[]{{\includegraphics[width=0.9\textwidth]{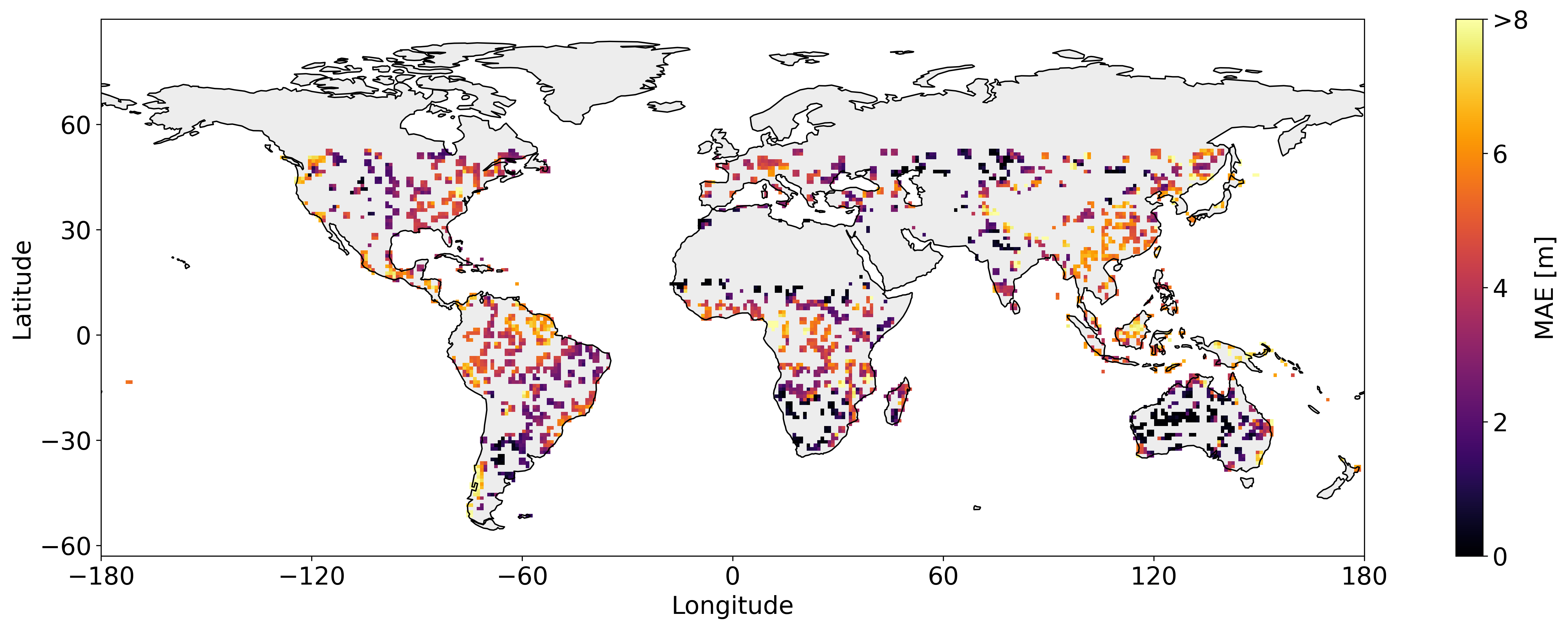} }}%
    
    \subfloat[]{{\includegraphics[width=0.9\textwidth]{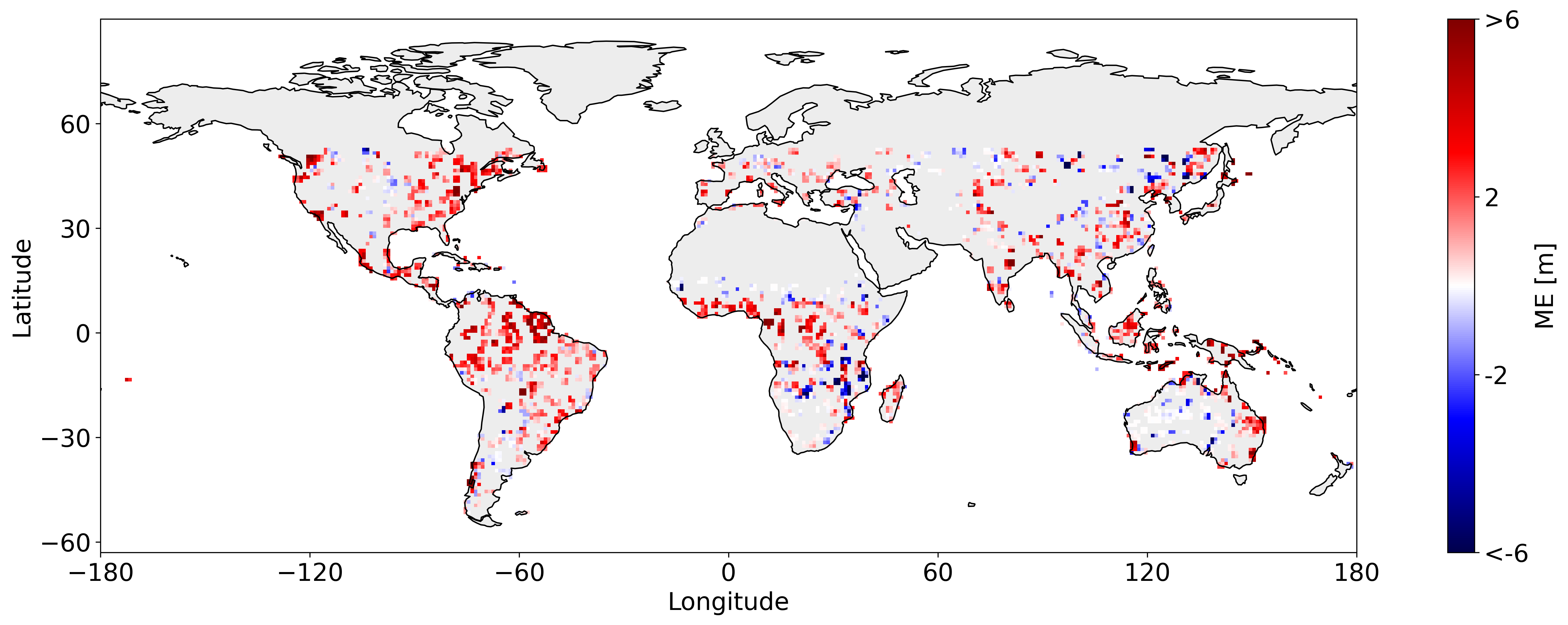} }}%
    
    \caption{Geographical error analysis on held-out GEDI validation data at 1 degree resolution ($\approx$111$\,$km on the equator). a) Root mean square error (RMSE). b) Mean absolute error (MAE). c) Mean error (ME), where negative ME mean an underestimation bias when the predictions are lower than the reference values.}
    \label{fig:map_errors}
\end{figure*}

\begin{figure*}
    \centering
    \subfloat[]{{\includegraphics[width=0.8\textwidth]{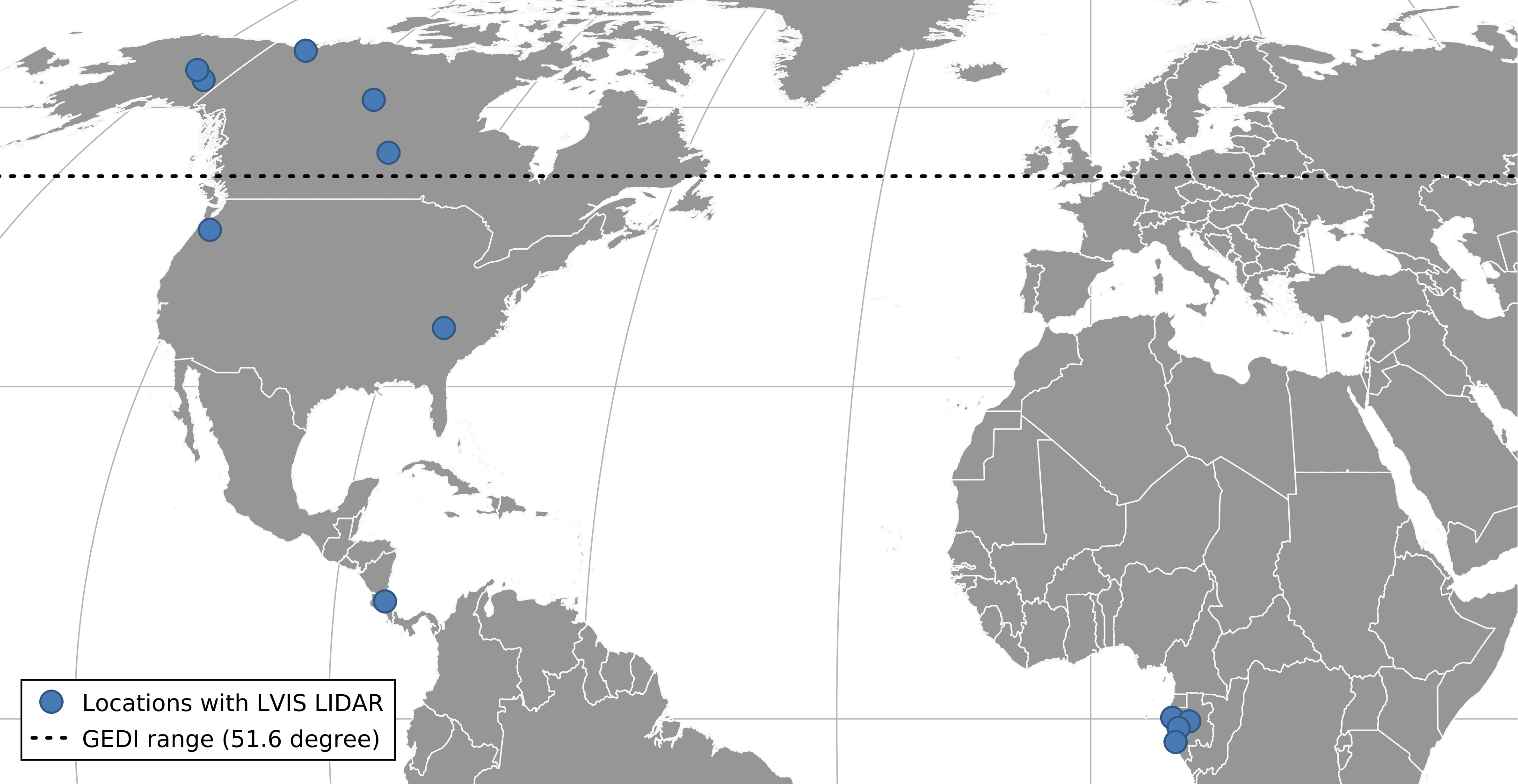} }}%
    
    \subfloat[]{{\includegraphics[width=0.33\textwidth]{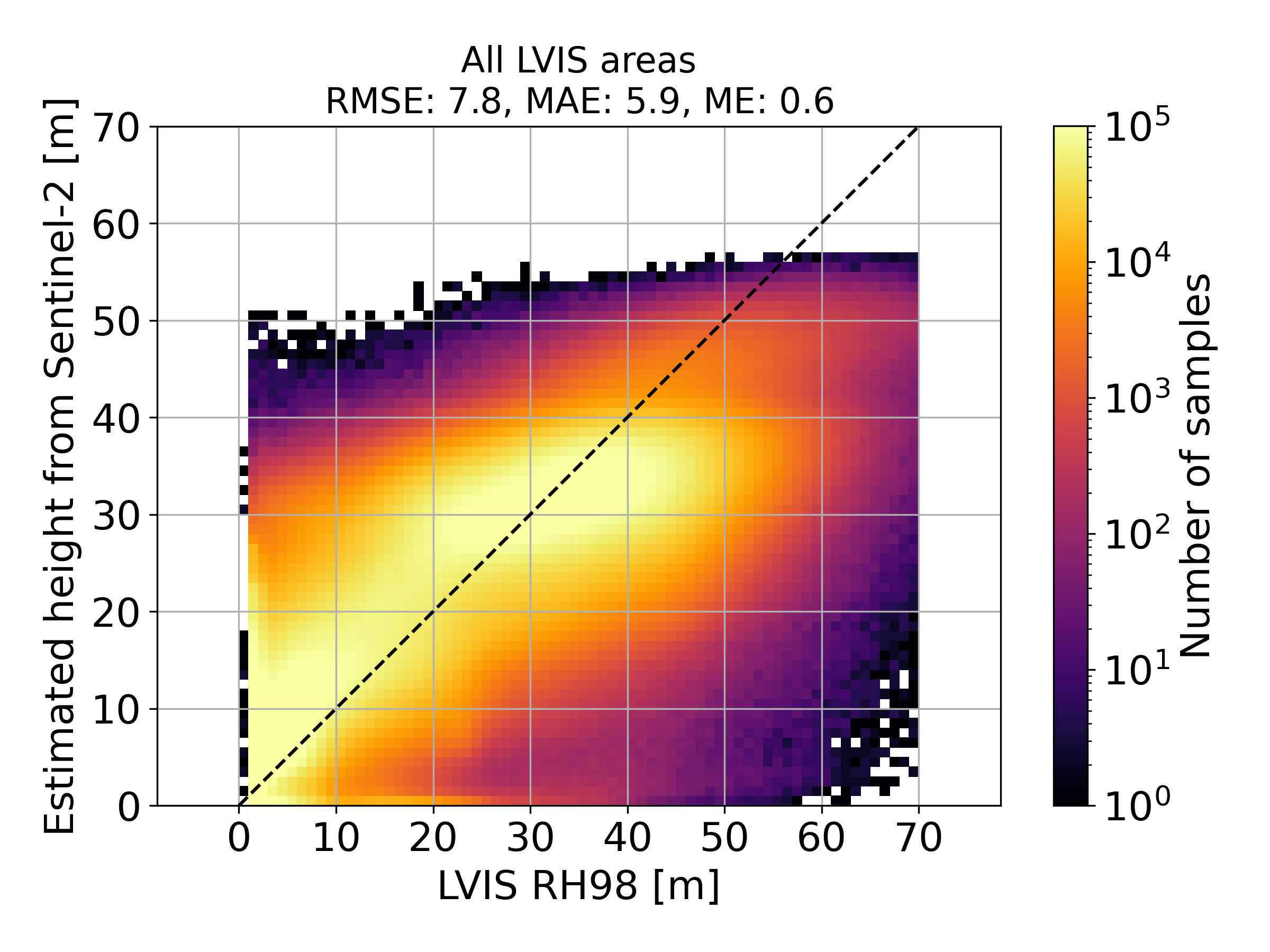} }}%
    \subfloat[]{{\includegraphics[width=0.33\textwidth]{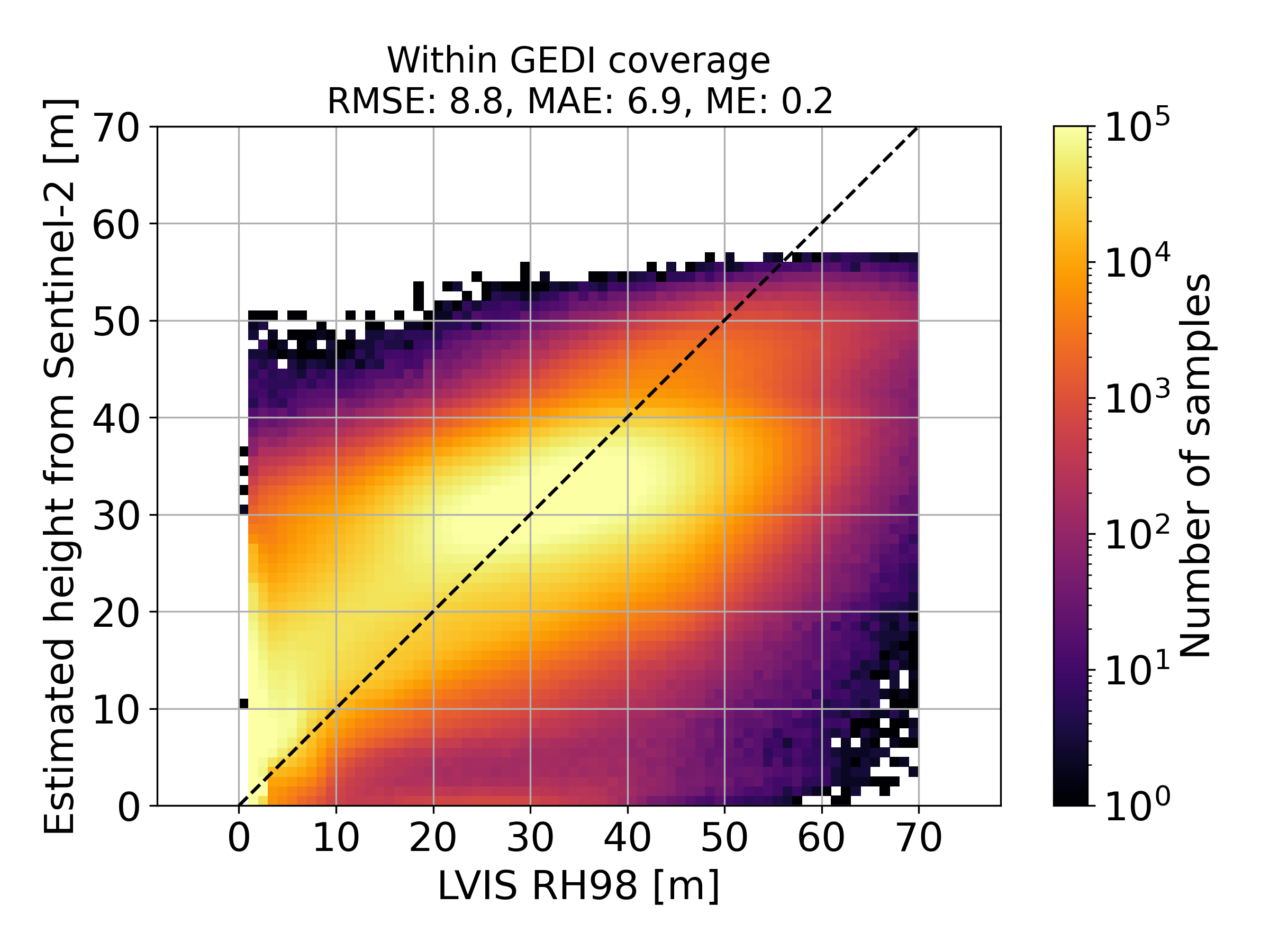} }}%
    \subfloat[]{{\includegraphics[width=0.33\textwidth]{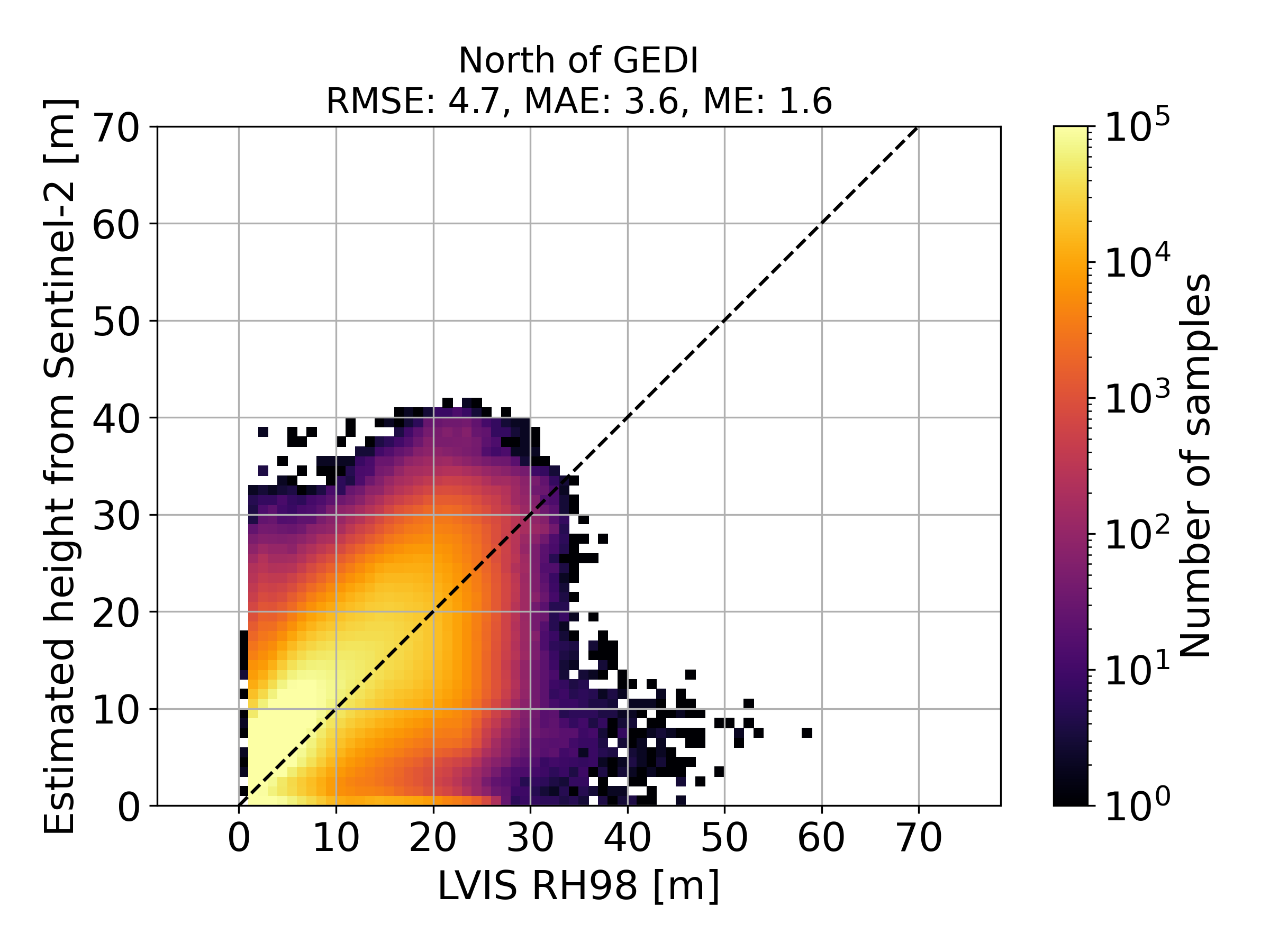} }}%
    
    \caption{Evaluation w.r.t.\ canopy top height (RH98) derived from independent LVIS airborne LIDAR data \citep{blair2004processing}. a) Locations of LVIS LIDAR campaigns. a-c) Confusion plots showing the relationship between LVIS reference data and predictions from Sentinel-2 for a) All available LVIS areas, b) Only regions within the GEDI range, and c) Only regions north of GEDI.}
    \label{fig:lvis_eval}
\end{figure*}

\begin{figure*}
    \centering
    \includegraphics[width=0.5\textwidth]{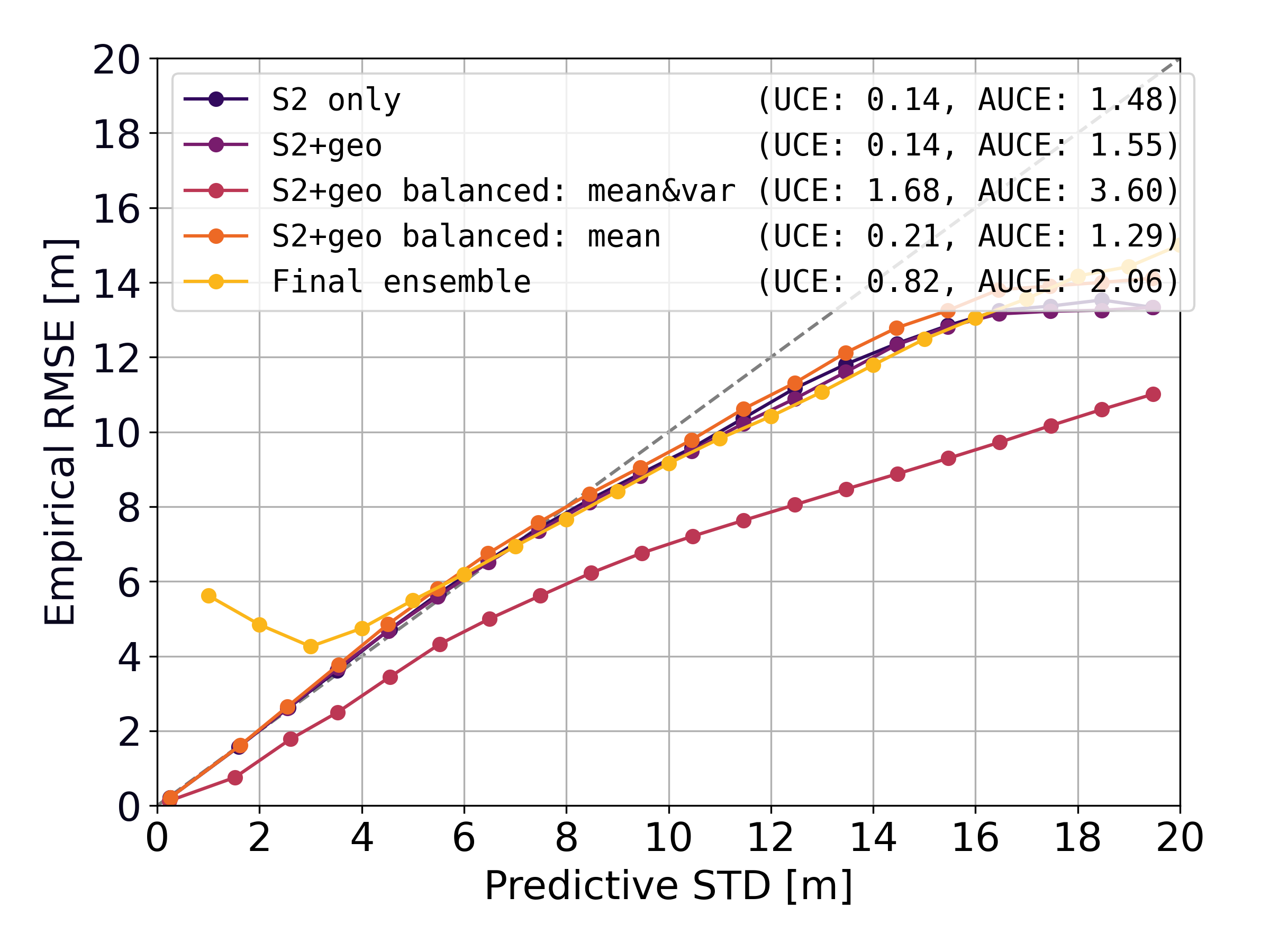}
    \caption{Calibration plot showing the relationship between the estimated predictive uncertainty and the empirical error.}
    \label{fig:calibration}
\end{figure*}

\begin{figure*}
    \centering
    \includegraphics[width=0.5\textwidth]{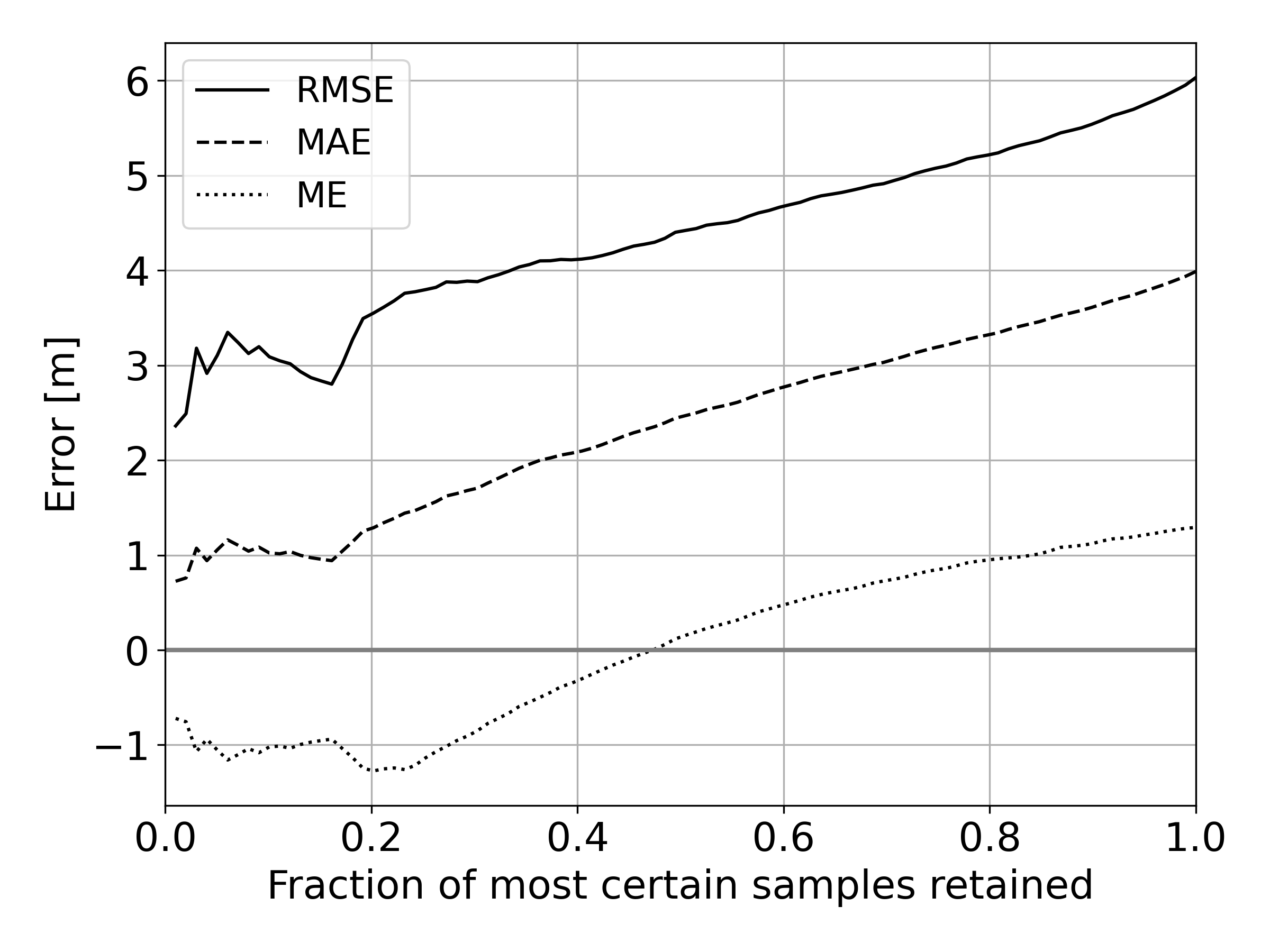}
    \caption{Improvement of overall error metrics when filtering out the most uncertain canopy height predictions with the help of the estimated predictive uncertainty.}
    \label{fig:filtering}
\end{figure*}

\begin{figure*}
    \centering
    \includegraphics[width=1.0\textwidth]{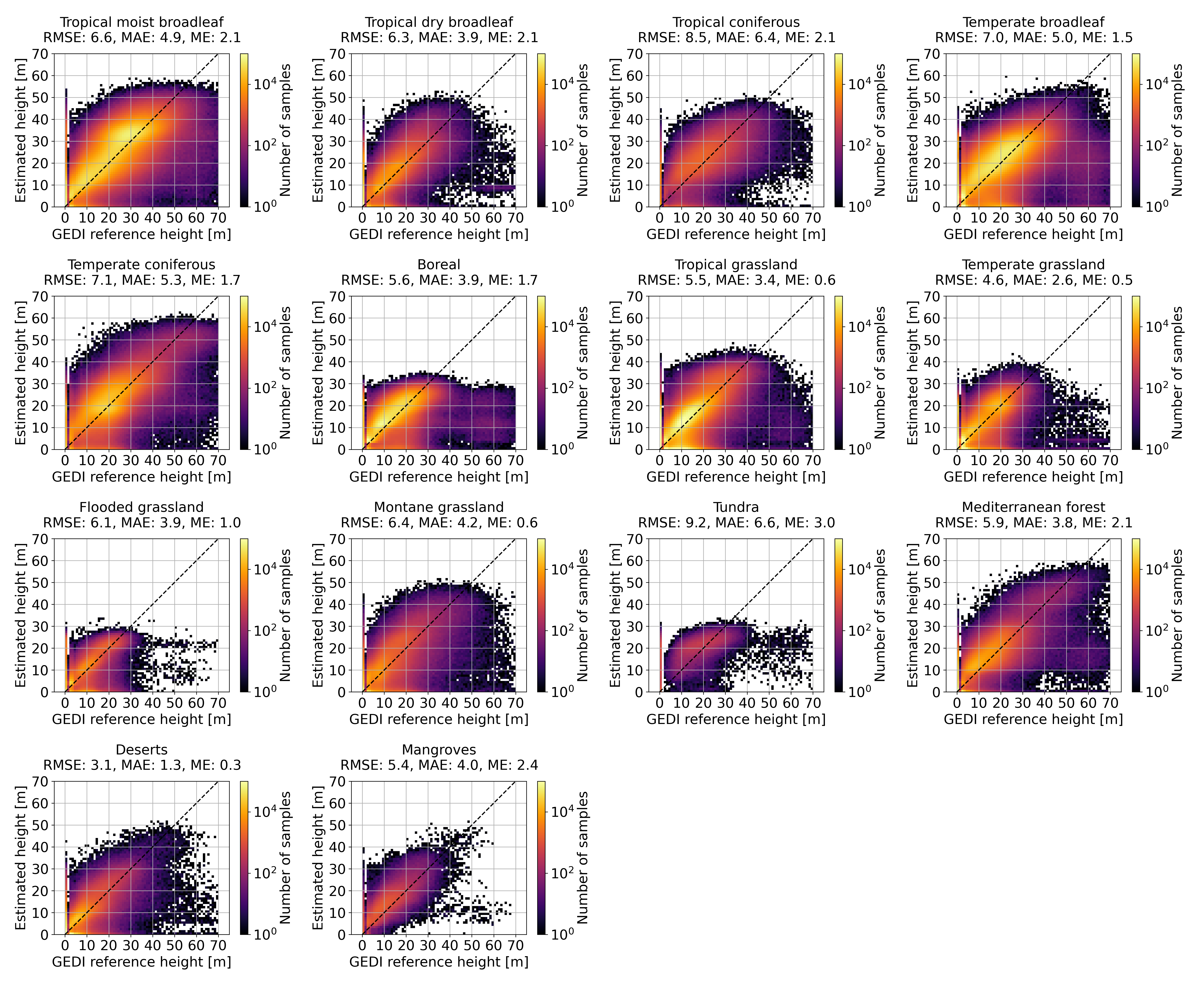}
    \caption{Biome-level confusion plots, showing the relationship between GEDI reference data and predictions from Sentinel-2.}
    \label{fig:biome_confusion}
\end{figure*}

\begin{figure*}
    \centering
    \includegraphics[width=1.0\textwidth]{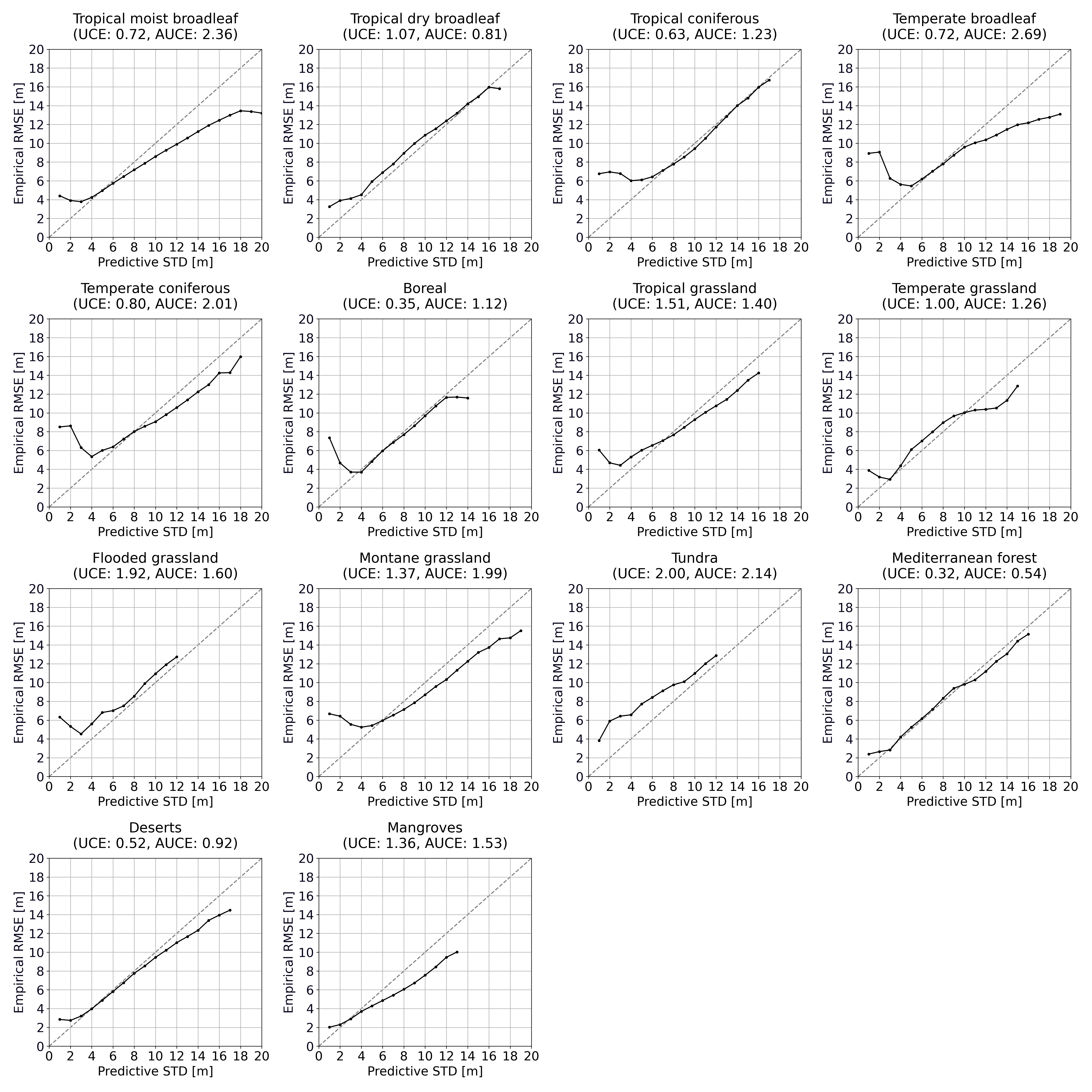}
    \caption{Biome-level calibration plots, showing the relationship between the estimated predictive uncertainty and the empirical error.}
    \label{fig:biome_calibration}
\end{figure*}

\begin{figure*}
    \centering
    \subfloat[]{{\includegraphics[width=0.5\textwidth]{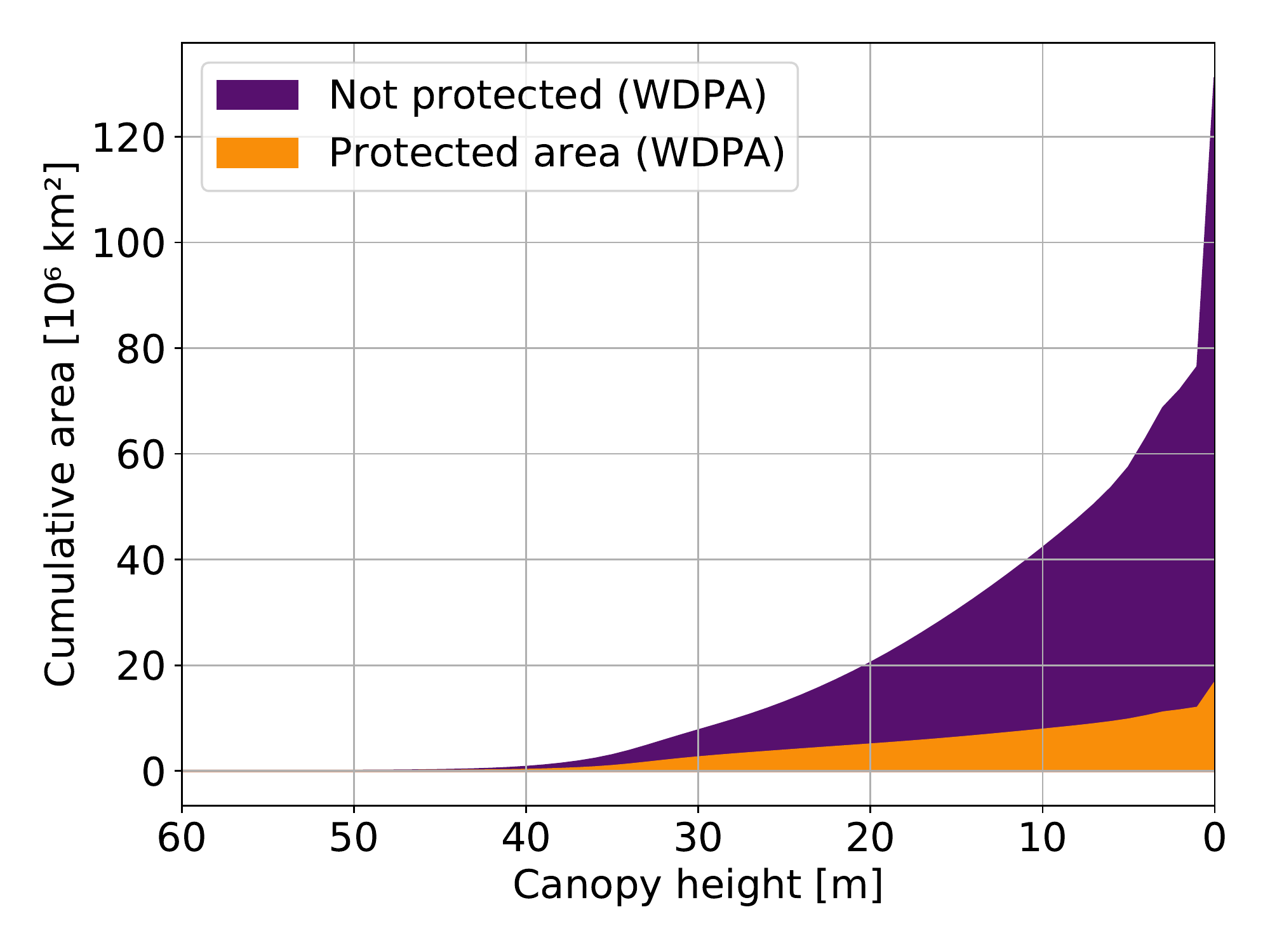} }}%
    
    \subfloat[]{{\includegraphics[width=1.0\textwidth]{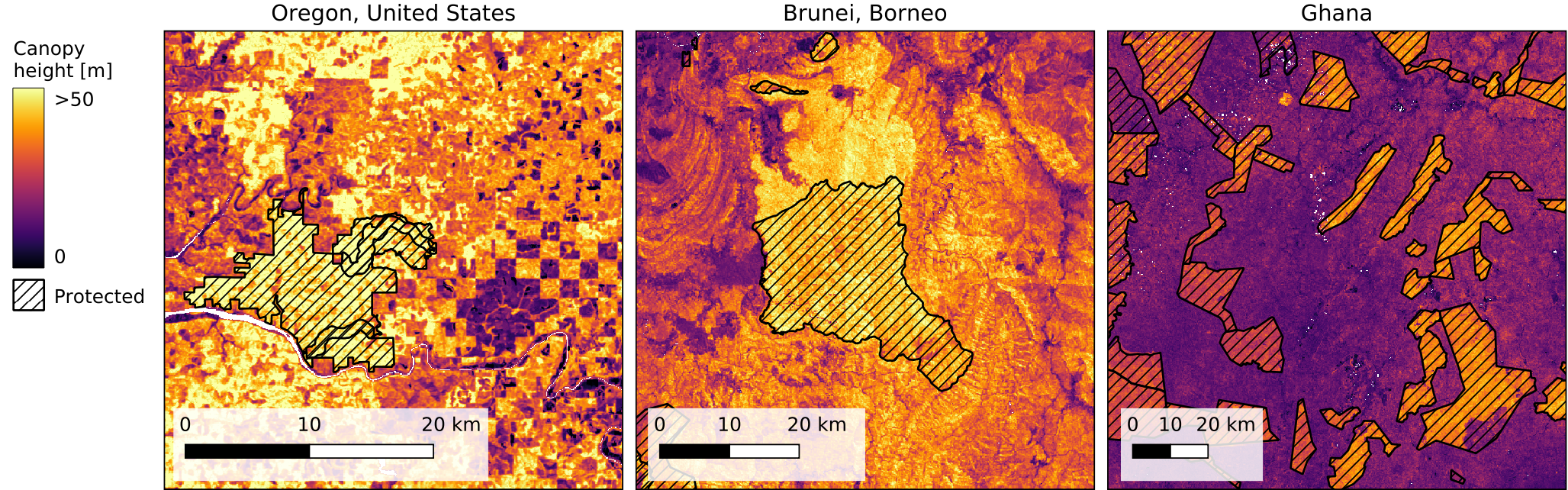} }}%
    
    \caption{Protected area analysis according to WDPA \citep{protectedplanet2021}. 
    a) Cumulative area covered by vegetation above a given height in protected areas and unprotected areas. The sum at height 0 equals the area of the global landmass (excluding Antarctica). 
    b) Examples where the dense canopy height map reveals the spatial patterns of protected areas. Left: "Devil's Staircase Wilderness" containing federally protected old-growth forest stands in the Oregon Coast Range. Center: "Ulu Temburong National Park" in Brunei, Borneo, established in 1991. Right: Protected areas in Ghana that indicate the strong impact of protection measures on the growing vegetation.}
    \label{fig:protected_areas}
\end{figure*}

\subsection{Comparison to UMD canopy height map}
\label{sec:comparison_umd_eth}
We compare our map against the UMD canopy height map, derived by fusing GEDI with Landsat image composites \citep{potapov2021mapping}.
As common reference, we using the independently created LVIS data (see Extended Data Table~\ref{tab:comparison_umd}). Since UMD data is not available north of 51.6$^\circ$  latitude, we use only LVIS regions within the GEDI coverage. While our map appears to be more accurate in all error metrics, the biggest difference is observed in the ME. The UMD map exhibits underestimation by -4.8$\,$m, likely due to saturation of tall canopies. Whereas our map has a negligible overestimation bias of 0.2$\,$m.

\begin{table*}[hb]
    \centering
    \begin{tabular}{@{}lrrr@{}}
    \toprule
               & RMSE & MAE & ME   \\ \midrule
    UMD \citep{potapov2021mapping}   & 9.6  & 7.4 & -4.8 \\ 
    ETH (ours) & 8.8  & 6.9 & 0.2  \\ \bottomrule \\
    \end{tabular}
    
    \caption{Performance comparison against UMD canopy height \citep{potapov2021mapping} with the canopy top height (RH98) derived from LVIS airborne LIDAR data \citep{blair2004processing}. Note that only LVIS data within the GEDI coverage is used, as the UMD data is not available north of 51.6 degrees Latitude.}
    \label{tab:comparison_umd}
\end{table*}

\end{document}